\documentclass[journal]{IEEEtran}
\usepackage{amsmath,amssymb,bm}
\usepackage{array,threeparttable,booktabs,multirow}
\usepackage{graphicx}
\usepackage[ruled]{algorithm2e}
\usepackage[noend]{algpseudocode}
\usepackage{cite}
\usepackage[dvipsnames]{xcolor}
\usepackage{setspace}
\usepackage[colorlinks=true, allcolors=blue]{hyperref}
\usepackage{pifont}
\usepackage{makecell}
\usepackage{diagbox}
\usepackage{setspace}
\usepackage{float}   
\usepackage{subfigure} 
\usepackage{enumitem}
\usepackage{orcidlink}

\begin{document}

\title{Deep Learning Advancements in Anomaly Detection: A Comprehensive Survey}

\author{Haoqi Huang, 
Ping Wang\orcidlink{0000-0002-1599-5480},~\IEEEmembership{Fellow,~IEEE}, 
Jianhua Pei\orcidlink{0000-0002-4066-9230},~\IEEEmembership{Graduate~Student~Member,~IEEE}, 
Jiacheng Wang\orcidlink{0000-0003-1252-8761}, 
Shahen Alexanian, 
and~Dusit~Niyato\orcidlink{0000-0002-7442-7416},~\IEEEmembership{Fellow,~IEEE}
\thanks{H. Huang, P. Wang and S. Alexanian are with the Lassonde School of Engineering, York University, Toronto, ON M3J 1P3, Canada (e-mail:joycehhq@yorku.ca; pingw@yorku.ca; yu263319@my.yorku.ca).

J. Pei is with the State Key Laboratory of Advanced Electromagnetic Technology, School of Electrical and Electronic Engineering, Huazhong University of Science and Technology, Wuhan 430074, China (e-mail: jianhuapei@hust.edu.cn).

J. Wang and D. Niyato are with the School of Computer Science and Engineering, Nanyang Technological University, Singapore (e-mail: jcwang\_cq@foxmail.com; dniyato@ntu.edu.sg).}
}

\markboth{IEEE Internet of Things Journal,~Vol.~x, No.~x, xxx~2025}%
{Shell \MakeLowercase{\textit{et al.}}: Bare Demo of IEEEtran.cls for IEEE Journals}

\maketitle

\begin{abstract}
The rapid expansion of data from diverse sources has made anomaly detection (AD) increasingly essential for identifying unexpected observations that may signal system failures, security breaches, or fraud. As datasets become more complex and high-dimensional, traditional detection methods struggle to effectively capture intricate patterns. Advances in deep learning have made AD methods more powerful and adaptable, improving their ability to handle high-dimensional and unstructured data. This survey provides a comprehensive review of over 180 recent studies, focusing on deep learning-based AD techniques. We categorize and analyze these methods into reconstruction-based and prediction-based approaches, highlighting their effectiveness in modeling complex data distributions. Additionally, we explore the integration of traditional and deep learning methods, highlighting how hybrid approaches combine the interpretability of traditional techniques with the flexibility of deep learning to enhance detection accuracy and model transparency. Finally, we identify open issues and propose future research directions to advance the field of AD. This review bridges gaps in existing literature and serves as a valuable resource for researchers and practitioners seeking to enhance AD techniques using deep learning.
\end{abstract}


\section{Introduction} \label{sec:intro}
\IEEEPARstart{A}{n} anomaly refers to an observation that significantly deviates from the expected behavior in a system, often appearing unusual, inconsistent, or unexpected \cite{1}. Despite the fact that outliers typically constitute only a small fraction of a dataset, they are often highly crucial because they carry important information and can reveal critical insights during analysis. Consequently, anomaly detection (AD) is the process of identifying such anomalous observations using various methods and algorithms, which aids decision-makers in better understanding data patterns and behaviors. 

The rapid development of the Internet of Things (IoT) has revolutionized the way data is generated, collected, and analyzed across various domains. IoT systems leverage a wide array of interconnected sensors and devices to collect massive amounts of real-time data in diverse applications, including smart cities \cite{194}, industrial automation \cite{195}, healthcare \cite{160}, and transportation \cite{197}, etc. This proliferation of sensor data introduces unprecedented opportunities for enhancing operational efficiency and decision-making processes. However, it also presents significant challenges, as the data is often high-dimensional, noisy, and prone to anomalies caused by faulty sensors, environmental changes, or malicious attacks \cite{115}. Detecting anomalies in data is critical for ensuring system reliability, security, and performance \cite{132}.

AD methodologies can be systematically classified according to various criteria. One prominent classification framework differentiates these methods into supervised, semi-supervised, and unsupervised approaches, predicated on the availability and nature of labeled data \cite{129}. Supervised learning-based AD algorithms necessitate a fully labeled dataset, where each data point is explicitly annotated as either normal or anomalous. This labeling process facilitates the model’s ability to discern and learn the underlying characteristics that differentiate anomalous instances from normal ones, thereby enhancing its detection accuracy. Semi-supervised learning-based methods, on the other hand, operate with a dataset comprising a substantial volume of unlabeled data alongside a smaller subset of labeled instances. These labels may include both normal and anomalous data, or in certain cases, solely normal instances \cite{16}. In scenarios where only normal data is labeled, the semi-supervised approach converges towards unsupervised methodologies, as the model predominantly learns normal behavior patterns and identifies anomalies as deviations from these learned patterns. Unsupervised learning-based AD methods eschew the need for labeled data entirely, leveraging the intrinsic structural properties of the dataset to autonomously identify anomalies \cite{120} \cite{125}. In practical applications, a significant portion of contemporary AD research gravitates towards unsupervised methods \cite{15}. This preference is largely driven by the substantial imbalance between the number of normal instances and anomalies, which complicates the acquisition of a sufficiently large labeled dataset required for effective supervised learning \cite{100}. Moreover, anomalies are frequently correlated with critical failures or hazardous events, rendering the labeling process both costly and logistically challenging. Another key classification criterion is the nature of the dataset, particularly whether it comprises time-series data, which distinguishes AD methods into time-series \cite{18} and non-temporal approaches. The applications of time-series and non-temporal AD will be discussed in detail in Section \ref{sec:related}.

In addition to the temporal aspect, AD techniques can also be categorized based on their underlying paradigms: traditional methods and deep learning-based methods. Traditional techniques encompass statistical approaches \cite{134},  distance-based methods \cite{sarmadi2020novel}, and clustering algorithms \cite{23}. These approaches generally rely on estimating the probability distribution of normal data to predict anomalies. However, since the early 20th century, the fields of data science, machine learning, deep learning, and artificial intelligence have witnessed exponential growth, with significant implications for AD \cite{2}. Particularly in recent years, the advent of soft-computing techniques has significantly influenced the development of deep learning-based methods. These techniques are characterized by their ability to handle imprecise, uncertain, and nonlinear data, making them highly suitable for applications involving deep learning. Consequently, deep learning-based methods have been propelled to the forefront due to their superior capability to learn expressive representations of complex data, including high-dimensional, temporal, spatial, and graph-structured data \cite{3}. By proficiently modeling intricate patterns and relationships inherent in the data, deep learning approaches have proven remarkably effective in identifying anomalies across a wide range of challenging and complex datasets. This paper concentrates specifically on AD methods based on deep learning models, with the objective of providing a comprehensive review of this rapidly evolving field.


\begin{figure}[!t]
\centering
\includegraphics[width=0.95\linewidth]{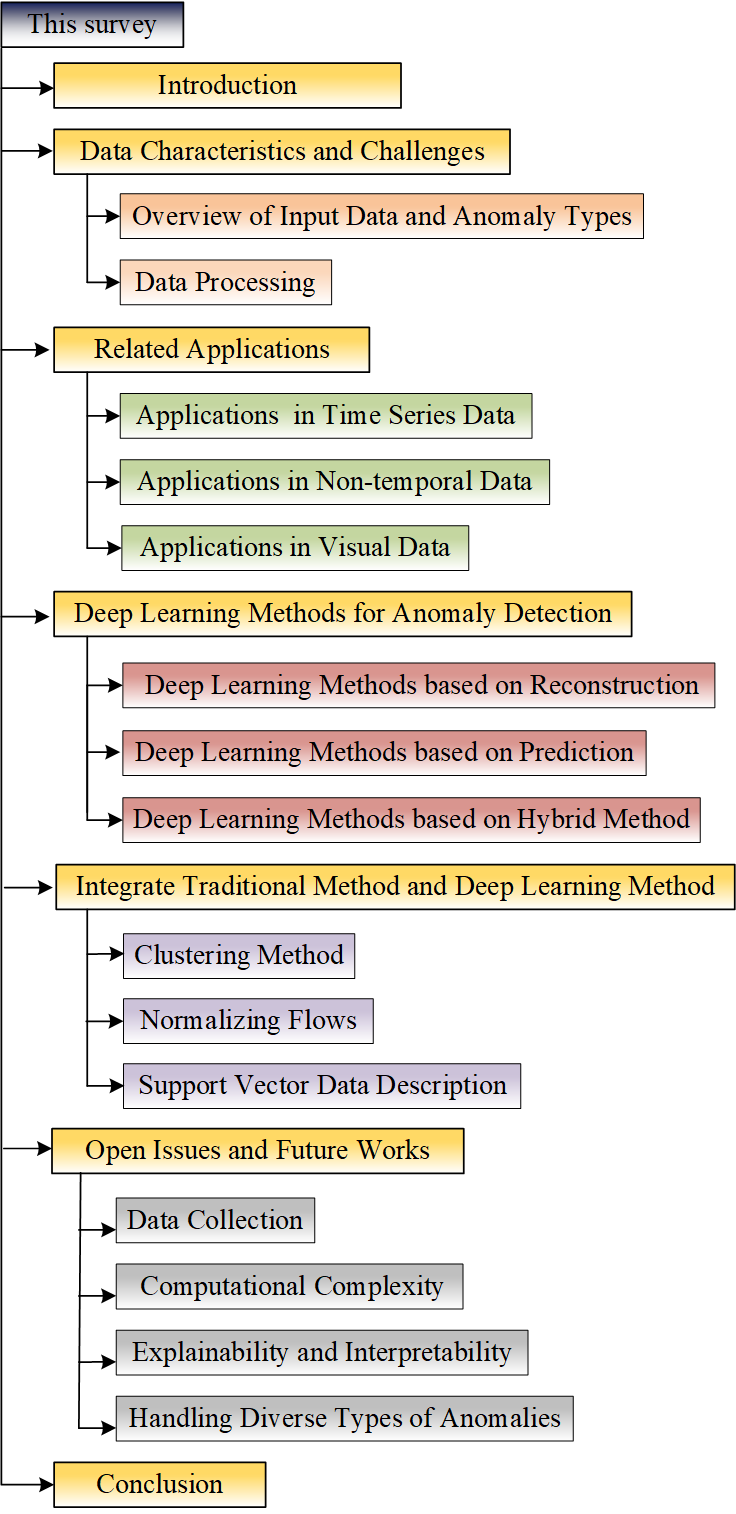}
\caption{The anatomy of this survey.}
\label{structure}
\end{figure}

\subsection{Contrasting Traditional Models with Deep Learning Models}
Traditional AD methods \cite{93}, such as statistical techniques, clustering algorithms \cite{24}, and Principal Component Analysis (PCA) \cite{135}, have long been established as reliable tools across a wide spectrum of applications due to their simplicity, interpretability, and low computational overhead. These characteristics make them particularly promising in scenarios where model transparency and efficiency are paramount. Statistical techniques, for example, provide clear, rule-based mechanisms for detecting anomalies, while clustering algorithms are effective in grouping similar data points and isolating outliers in relatively low-dimensional datasets. Similarly, PCA has been widely adopted for dimensionality reduction, enabling effective AD by isolating principal components that capture major variations in the data \cite{23}. Despite these advantages, traditional methods often encounter significant limitations when applied to modern, complex datasets. Statistical techniques generally assume that data adheres to specific distributions. However, this assumption is rarely met in real-world scenarios, where data often exhibits non-Gaussian distributions and heavy tails. Clustering-based methods, while useful in many contexts,  check to accurately define clusters, particularly when anomalies do not present clear separability from normal data. PCA, on the other hand, relies heavily on the assumption of linearity and extensive feature engineering, making it less effective at capturing the nuanced, non-linear patterns prevalent in high-dimensional datasets \cite{135}. These constraints have prompted a shift towards more advanced approaches capable of handling the increasing complexity of modern data.

In contrast, deep learning models have recently emerged as a powerful alternative, addressing many of the shortcomings inherent in traditional approaches. Deep neural networks (DNNs) possess the capacity to autonomously learn complex patterns and hierarchical representations from raw data, thereby obviating the need for labor-intensive feature engineering \cite{90}. This characteristic is particularly advantageous in the detection of subtle and multifaceted anomalies that might elude traditional methods \cite{8}. By leveraging their multi-layered architectures, deep learning models excel in processing high-dimensional and unstructured data, such as images, videos, and text, which are often challenging for conventional methods to handle effectively \cite{128}. These models are adept at capturing non-linear relationships and interactions within the data, offering a more flexible and robust framework for AD \cite{32}. Consequently, there has been a significant shift away from purely traditional AD techniques towards the adoption of deep learning methodologies.

Nonetheless, it is crucial to acknowledge that traditional AD models retain certain advantages, notably in their simplicity, interpretability, and lower computational overheads \cite{104}. These characteristics make them particularly appealing in scenarios where model transparency and computational efficiency are crucial. In recognition of these strengths, Section V of this paper will introduce and discuss various existing approaches that integrate traditional methods with deep learning techniques. These hybrid methods aim to leverage the strengths of both paradigms, resulting in more robust and efficient AD systems. 
\subsection{Comparison With Existing Surveys}
In recent years, the field of AD has seen a surge in research, particularly with the advent of deep learning methods. Numerous surveys have been published, each attempting to provide a comprehensive overview of the field. However, many of these surveys focus on broader historical developments or cover deep learning techniques only up to a certain point in time. For example, surveys such as \cite{3}, \cite{4}, \cite{11}, and \cite{90} primarily cover techniques developed up to 2020. While these surveys are valuable, they do not reflect the most recent advancements in the field. Furthermore, specific models such as Generative Adversarial Network (GAN)-based AD have been explored in-depth by studies \cite{5}, \cite{75}, \cite{28}, \cite{76}, and \cite{148}.  However, these studies primarily address foundational approaches and lack coverage of advanced techniques like conditional GANs, cycle-consistent GANs, and GANs integrated with self-supervised learning. Emerging hybrid models, combining GANs with Variational Autoencoders (VAEs) or autoencoders for improved robustness, are also underrepresented. In contrast, our survey covers the literature from 2019 to 2024, providing a timely and comprehensive overview of the latest advancements. By focusing on recent trends and evolving techniques, including enhanced architectures and hybrid frameworks, our work offers a more current perspective, bridging existing gaps and guiding future research directions in AD.


\subsection{Contributions and Structure}
This survey systematically reviews over 160 recent research papers on AD, including publications from leading journals (IEEE, ACM, Springer, Elsevier) and top-tier conferences (AAAI, CCS, ICCV) spanning from 2019 to 2024. By focusing on cutting-edge advancements in deep learning-based methods, this survey ensures a comprehensive and up-to-date overview of the field. The contributions of this survey are summarized as follows:
\begin{itemize}
\item This survey addresses gaps in prior surveys by highlighting advanced techniques that were previously underexplored, including conditional GANs, cycle-consistent GANs, and hybrid frameworks combining GANs with VAEs. These models are introduced and analyzed to demonstrate their strengths and weaknesses.

\item This survey provides a detailed comparison of reconstruction-based and prediction-based methods. To enhance clarity and usability, we summarize key strengths, weaknesses, and applications in structured tables, offering readers insights into the trade-offs of different models.

\item Recognizing the strengths of traditional methods, this survey explores their integration with deep learning models. Hybrid approaches, such as clustering, normalizing flows, and support vector data descriptions combined with deep learning, are analyzed to address complex challenges in AD.
\end{itemize}

The organization of this survey is shown in Fig.\ref{structure}. Section \ref{sec:data} provides an overview of data characteristics and anomaly types, followed by a discussion of common data processing challenges and mitigation strategies critical to effective AD. Section \ref{sec:related} explores the related applications of AD. Section \ref{sec:deep} categorizes and analyzes deep learning methods for AD, highlighting their effectiveness and limitations. Section \ref{sec:integrate} discusses the integration of traditional methods with deep learning, including clustering methods, normalizing flows, and support vector data descriptions. Section \ref{sec:open} highlights open issues and future directions, such as challenges in data collection, computational complexity, explainability, and handling diverse anomaly types. Finally, Section \ref{sec:con} concludes the survey with a summary and potential directions for future research.

\section{Data Characteristics and Challenges} \label{sec:data}

\subsection{Overview of Input Data and Anomaly Types}
In AD, input data presents unique challenges due to its structure, dimensionality, and temporal nature. Different types of data require specialized techniques to effectively identify anomalies, and the nature of anomalies themselves can vary greatly depending on the domain and data format \cite{4}. For instance, visual data such as images and videos may exhibit anomalies associated with spatial or temporal inconsistencies, while time series data often involves anomalies related to trends or sudden changes in values over time. To better understand these variations, we first categorize data into textual, audio, image, and video formats, highlighting their respective characteristics and the challenges they pose for AD. Beyond this classification, data can also be viewed through the lens of temporal dependencies, distinguishing between time-series data, which captures sequential patterns over time, and non-temporal data, where observations are independent of temporal order. This dual perspective provides a comprehensive framework for analyzing how different types of anomalies manifest across various data formats. Furthermore, the nature of anomalies themselves can vary depending on the data format. Point anomalies, sequence anomalies, and outliers may all manifest differently across different data types and structures. Understanding these distinctions is essential for selecting the appropriate AD techniques \cite{11}, as a deep understanding of data characteristics and anomaly types ensures that detection methods are effectively tailored to capture the specific behaviors and patterns indicative of anomalies.

\subsubsection{Categorization by Data Type}
\begin{itemize}
\item \textbf{Textual Data:}
Textual data consists of sequences of discrete symbols, such as characters, words, or phrases, structured in a linear format. Unlike other data types, textual data conveys information through syntactic and semantic relationships. It can be found in various forms, including documents, chat messages, emails, and system logs. Anomalies in textual data may appear as irregular word sequences, syntactic inconsistencies, missing or misplaced words, or semantically incoherent phrases.
\item \textbf{Audio Data:}
Audio data captures variations in amplitude and frequency over time, representing spoken language, environmental sounds, or machine signals. It can be stored as waveforms or transformed into frequency-domain representations like spectrograms. Unlike textual data, audio data is continuous and often requires spectral analysis to extract meaningful patterns. Anomalies in audio data manifest as unexpected distortions, unusual frequency shifts, missing segments, or abnormal sound patterns caused by malfunctioning equipment, altered speech, or environmental noise.
\item \textbf{Image Data:}
Image data consists of two-dimensional pixel grids, where each pixel represents intensity or color information. Unlike sequential data, image data encodes spatial relationships, capturing textures, shapes, and patterns. Image anomalies often appear as distortions, irregular textures, missing components, or unexpected objects that deviate from normal patterns. For instance, these can result from manufacturing defects, medical imaging errors, or environmental changes in satellite imagery.
\item \textbf{Video Data:}
Video data extends image data by incorporating a temporal dimension, forming sequences of frames over time. Each frame within a video is an image, and the relationships between frames capture motion and dynamic interactions \cite{9}. Unlike static images, video data requires modeling temporal dependencies, making AD more complex. Anomalies in video data include irregular movements, unexpected scene transitions, or unusual object behaviors, which are commonly observed in surveillance footage, traffic monitoring, and activity recognition.
\item \textbf{Tabular Data:}
Tabular data consists of structured records organized in rows and columns, where each row represents an entity or event, and each column corresponds to an attribute. This type of data is widely used in databases, spreadsheets, financial records, and sensor logs. Unlike the other data types, tabular data can contain numerical, categorical, or mixed-format information. Anomalies in tabular data include missing values, unexpected categorical labels, numerical outliers, or inconsistent relationships between attributes.
\end{itemize}
\subsubsection{Categorization by Temporal Characteristics}
\begin{itemize}
    \item \textbf{Time-based data:}
Time-based data can be represented as a sequence of observations recorded over time, and it may consist of either a single variable (univariate) or multiple variables (multivariate). We can generalize the representation of both univariate and multivariate time series using the following formula:$X = \{ x_{t,j} \}_{t \in T, j \in J}$,
where \( t \in T \) denotes the time index, with \( t \) representing a specific time step and \( T \) being the set of all time steps in the dataset. Similarly, \( j \in J \) represents the dimension or variable index, where \( j \) refers to a particular variable and \( J \) is the set of all variables or dimensions in the data. When \( \left| J \right| = 1 \), the series is univariate, meaning there is only one variable observed over time. In contrast, when \( \left| J \right| > 1 \), the series is multivariate, indicating that multiple variables are recorded simultaneously at each time step. Each observation \( x_{t,j} \) corresponds to the value of the \( j \)-th variable at time \( t \). Among the five previously introduced data types, audio, video, and certain types of textual and tabular data are inherently time-based. Audio data is naturally sequential, with sound signals evolving over time, making anomalies such as distortions or frequency shifts dependent on temporal patterns. Video data extends image sequences over time, requiring the detection of abnormal object movements, scene transitions, or motion inconsistencies. Textual data, such as streaming logs, system event records, or chat conversations, also exhibits temporal dependencies, where anomalies may appear as unexpected event sequences or irregular timing between log entries. Similarly, tabular data in the form of financial transactions, sensor readings, or stock prices follows a time-series format, where anomalies may indicate fraud, equipment failure, or unusual market behaviors.
\item \textbf{Non-temporal data:}
Non-temporal data refers to observations that lack a temporal sequence, where the relationships between data points are independent of time. Such data is prevalent across industries that rely on static datasets or event-based observations. AD in non-temporal data focuses on identifying irregularities by analyzing data characteristics, patterns, or statistical properties rather than temporal dependencies. This process is crucial for uncovering hidden risks, fraudulent activities, or system malfunctions in contexts where time is not a defining factor. Among the five data types, image and certain types of tabular data are the most common forms of non-temporal data. Image data, such as medical scans, industrial defect detection images, or satellite photos, captures spatial relationships but does not depend on a temporal sequence. Anomalies in such data typically appear as unusual textures, distortions, or unexpected objects. Tabular data, when not used for time-series analysis, is also non-temporal, such as customer records, product attributes, or static financial datasets. In these cases, AD focuses on identifying outliers, inconsistencies, or unusual relationships between different features rather than changes over time.
\end{itemize}
\subsubsection{Types of Anomalies}
\begin{itemize}
    \item \textbf{Point Anomalies}: 
    A single data point deviates significantly from the expected behavior in the dataset. These are common across both time-based and non-time-based data, representing sudden outliers or unusual values.

    \item \textbf{Contextual Anomalies}: 
    A data point is considered anomalous only when it is analyzed within a specific context or surrounding data. In time-based data, this could involve seasonal trends or time-of-day variations, whereas in non-time-based data, it could depend on relationships between variables.

    \item \textbf{Subsequence Anomalies}: 
    A contiguous sequence of data points behaves abnormally, typically found in time series data. These anomalies are significant when the temporal order of data points plays a key role in detecting deviations from expected patterns.

    \item \textbf{Cluster-based and Correlation Anomalies}: 
    Anomalies that occur when a group of data points, or relationships between variables, deviate from expected patterns. This is more prominent in non-time-based data, where detecting irregular clusters or correlations between features is essential for AD.
\end{itemize}
\subsection{Data Processing}
Effective AD requires careful preparation and preprocessing of input data to ensure that detection algorithms can operate effectively. In many cases, raw data contains inherent challenges that can significantly hinder the performance of AD models. These challenges arise from the complexity of real-world data, including high dimensionality, missing or sparse values, skewed class distributions, and noise that can obscure true anomalies. Without addressing these issues, AD methods may struggle to accurately identify rare or subtle deviations in the data, leading to false positives, missed anomalies, or inefficient computations. Therefore, appropriate data preprocessing steps are crucial for improving detection accuracy, robustness, and overall system reliability. This subsection outlines some of the most common data processing issues and their implications for AD, along with strategies to mitigate these challenges.
\subsubsection{Dimensionality}
High-dimensional data makes AD more complex due to the ``curse of dimensionality". As datasets expand in size and complexity—particularly with the rise of ``big data", characterized by large-scale, high-velocity data generated from diverse sources, it becomes increasingly difficult for AD methods to maintain accuracy \cite{91}.  To address this issue, dimensionality reduction is a common approach that transforms a large set of input features into a smaller, more focused feature set \cite{170}. While traditional methods such as PCA \cite{172} are frequently used, they may struggle to capture non-linear relationships in complex data. For instance, Sakurada \textit{et al.} \cite{171} compare autoencoders, which perform non-linear dimensionality reduction, with linear PCA and kernel PCA on both synthetic and real-world datasets. The study reveals that on the nonlinear and high-dimensional synthetic Lorenz dataset, AE achieved a relative AUC improvement of 26.83\% compared to linear PCA. This highlights that autoencoders can even detect anomalies in data with relatively high intrinsic dimensionality, where linear PCA struggles to perform.
\subsubsection{Sparsity}
Sparse data, where many values are missing or incomplete, poses significant challenges for AD. Sparse datasets can lead to reduced detection accuracy, as missing or incomplete data points may obscure the underlying patterns necessary for detecting anomalies \cite{91}. Cheng \textit{et al.} \cite{174} highlight that in high-dimensional settings, the sparsity problem is further amplified as the data becomes more spread out, increasing the risk of missing critical information that signals anomalies. To address these challenges, Li \textit{et al.} \cite{173} propose an improved low-rank and sparse decomposition model (LSDM) for hyperspectral AD. Their approach models sparse components as a Gaussian Mixture (MoG), effectively capturing anomalous patterns within complex datasets by leveraging the low-rank structure. In contrast, Han \textit{et al.} \cite{175} take a different approach by introducing sparse autoencoders to learn sparse latent representations from high-dimensional input data. Through experiments on three real-world cyber-physical system datasets, the study shows that mining sparse latent patterns from high-dimensional time series can significantly improve the robustness of AD models.
\subsubsection{Class Imbalance}
In most AD tasks, the occurrence of anomalies is significantly rarer than normal data points, resulting in a class imbalance problem. This imbalance can cause detection algorithms to be overly biased toward the majority class (normal data), leading to a higher rate of false negatives where critical anomalies are missed. In imbalanced datasets, it is often possible to achieve an overall high accuracy, while the recall score for the minority class (anomalies) remains very low \cite{176}. Traditional methods to mitigate this issue involve oversampling the minority class or undersampling the majority class \cite{178}. Recent research has increasingly focused on introducing Data Generation Models (DGM) to improve the representation of the minority class in AD. For instance, Dlamini \textit{et al.} \cite{177} use Conditional Generative Adversarial Networks (CGANs) to generate synthetic samples for the minority class and combines this with KL divergence to guide the model in accurately learning the distribution of the minority class.
\subsubsection{Noise in Data}
Noise refers to random or irrelevant information present in the data, which can obscure true anomalies and lead to false positives. In addition, during the training process of AD models, the high complexity of the model and the presence of noisy data can lead to overfitting, where the model inadvertently learns to fit the reconstruction error from noisy inputs rather than focusing on genuine anomalies \cite{179}. To reduce the impact of noisy data, Zhang \textit{et al.} \cite{180} incorporate a Maximum Mean Discrepancy (MMD) to encourage the distribution of low-dimensional representations to approximate a target distribution. The goal is to align the distribution of noisy data with that of normal training data, thereby reducing the risk of overfitting. Furthermore, Chen \textit{et al.} \cite{181} propose a novel method called Noise Modulated Adversarial Learning, where noise images from a predefined normal distribution are fed into the discriminator network as negative samples. This adversarial process modulates the training of the reconstruction network, balancing the learning between the two networks to improve robustness against noise.
\subsubsection{Privacy of data}
In many fields, such as healthcare, finance, and cybersecurity, data used for AD often contains sensitive or personal information. Ensuring the privacy and security of this data is paramount, as improper handling could lead to serious legal and ethical violations. Hassan \textit{et al.} \cite{182} conducte an in-depth investigation into the privacy of AD models in blockchain technology. To address these privacy concerns, Federated Learning (FL), a distributed machine learning paradigm, has emerged as a promising supplement to AD \cite{156}. FL allows distributed clients to collaboratively train a shared model while protecting the privacy of their local data. For example, Idrissi \textit{et al.} \cite{40} propose Fed-ANIDS, which leverages FL to address the privacy issues associated with centralized Network Intrusion Detection Systems (NIDS). This model was applied to various settings and popular datasets, demonstrating its ability to achieve high performance while preserving the privacy of distributed client data. Cui \textit{et al.} \cite{183} further introduce GAN into FL and design a new algorithm model that injects controllable noise into local model parameters, ensuring both AD utility and compliance with differential privacy requirements. 

\section{Related Applications} \label{sec:related}
With the rapid advancement of deep learning models, AD has become more efficient and adaptable. These sophisticated models have been widely applied across various domains, enhancing the ability to identify irregular patterns in complex and high-dimensional datasets. In the previous chapter, we categorized data based on temporal characteristics into time-series and non-time-series data. However, visual data presents unique challenges, detection requirements, and a wide range of applications, making it difficult to be strictly classified as either time-series or non-time-series data. It can be static (e.g., images) or dynamic (e.g., videos), where images are typically considered non-time-series data, while videos fall under time-series data. Visual data is extensively used in fields such as medical imaging, autonomous systems, and surveillance, where detecting anomalies requires specialized deep learning techniques that differ from traditional numerical or categorical data analysis. To better reflect its broad applications and distinct computational needs, we discuss visual data separately. Based on this classification, we will now explore the applications of deep learning in AD from three perspectives: time-series data, non-temporal data, and visual data.
\subsection{Applications in Time Series Data}
Time series data, defined by its sequential nature over time, is fundamental to many systems where the temporal order of events critically influences analysis and decision-making processes. AD in time series data has become an indispensable technique across various industries, enabling the early detection of irregular patterns that may indicate underlying issues or emerging threats. The applications of time series AD are extensive, impacting critical areas such as traffic monitoring, power system management, and healthcare. In the following sections, we present how these applications leverage AD to enhance operational efficiency, ensure system reliability, and improve safety across these fields. 
\subsubsection{Traffic Monitoring}
Time series AD plays a pivotal role in modern traffic management systems. As demonstrated in \cite{139}, real-time data from loop detection sensors are integrated and analyzed to predict traffic volume and enhance system safety. The ability to detect anomalies in traffic patterns is essential for anticipating and responding to potential incidents before they escalate. For instance, Li \textit{et al.} \cite{142} present a method that identifies traffic incidents by detecting anomalies in traffic time series data, thereby helping users avoid accidents and reduce travel time. Furthermore, high-speed driving is identified as a significant contributor to traffic accidents \cite{20}. By monitoring and analyzing sudden increases in vehicle speed, AD techniques can predict and prevent accidents more effectively, providing a critical tool for improving road safety. Zhao \textit{et al.} \cite{10} further validate the efficacy of unsupervised AD methods in assessing elevated road traffic accident risks, specifically by analyzing volume and speed data from traffic on Yan’an elevated road. This approach enhances the ability to detect and respond to hazardous traffic conditions in real-time, underscoring the indispensable role of AD in traffic management.  
\subsubsection{Power System}
AD is a vital element in ensuring the stability, security, and reliability of electrical grids. By continuously monitoring grid data, these techniques can swiftly identify deviations from normal operational patterns, which may indicate issues such as natural faults or malicious cyber-attacks. The ability to detect these anomalies in real-time is crucial for preventing potential outages and maintaining a consistent power supply. For instance, Li \textit{et al.} \cite{143} highlight that accurate and real-time AD can enhance grid stability by over 20\%, providing rapid response capabilities that significantly bolster the system's defense against both natural disruptions and cyber threats. Furthermore, the introduction of a residential electrical load AD framework, as demonstrated in \cite{144}, has been shown to significantly improve both load prediction accuracy and AD, thereby optimizing demand-side management (DSM) in residential areas. In terms of cybersecurity, the MENSA Intrusion Detection System (IDS) \cite{153} has proven to be a formidable tool in smart grid environments, effectively detecting operational anomalies and classifying a wide range of cyberattacks. This capability not only protects critical infrastructure but also underscores the indispensable role of AD in modern power system management.
\subsubsection{Healthcare}
AD plays a crucial role in healthcare by enabling continuous monitoring of patient vital signs, such as heart rate and blood pressure, to swiftly identify abnormal conditions that may require urgent medical intervention. The application of AD in medical signal analysis is particularly important, as highlighted in \cite{154}, where the identification of data samples that deviate from the typical data distribution can reveal underlying issues such as noise, changes in a patient's condition, or the emergence of new and previously undetected medical conditions. This capability is essential for ensuring accurate diagnosis and timely patient care. Furthermore, Keeley \textit{et al.} \cite{118} demonstrate that AD algorithms can effectively identify irregularities in heart rate data, which not only facilitates faster emergency responses but also provides deeper insights into a patient's health status. This, in turn, enhances overall patient care while also reducing the cognitive load on healthcare professionals by automating the detection of potential issues.
\subsection{Applications in Non-temporal Data}
AD in non-temporal data plays a critical role in ensuring operational integrity, security, and financial stability. By focusing on identifying irregularities within independent events or static datasets, it addresses potential risks such as fraud, system failures, and malicious activities. Unlike time-series applications, non-temporal AD leverages data patterns and statistical analysis to uncover deviations that signal anomalies. In the following, we present specific applications across domains such as finance and cybersecurity, showcasing its significant impact on safeguarding critical systems and operations.
\subsubsection{Finance} 
In the financial sector, non-temporal data AD is pivotal for identifying fraudulent transactions, credit scoring anomalies, and unusual trading activities. Unlike time series data, these financial fraud detection tasks often involve independent events, such as individual transactions or credit score evaluations, which do not rely on temporal sequences. Instead, the focus is on transaction characteristics and patterns that may indicate fraudulent behavior. Various data mining techniques, including SVM, Naïve Bayes, and Random Forest, are extensively employed to detect different forms of financial fraud, such as bank fraud, insurance fraud, financial statement fraud, and cryptocurrency fraud \cite{146}. As highlighted by \cite{13}, AD is critical in quickly identifying activities that deviate from normal patterns, thereby enabling rapid intervention to minimize financial losses.
\subsubsection{Cybersecurity} AD is a fundamental component of maintaining a secure and resilient cyberspace. As \cite{157} points out, advanced security controls and resilience analysis are crucial during the early stages of system deployment to ensure long-term sustainability. AD plays a pivotal role in this process by identifying unauthorized access, malicious activities, and network intrusions that deviate from established norms. This capability is essential for safeguarding network security and preventing potential breaches. Early research in deep learning-based network intrusion detection focused on architectures such as Autoencoders (AE), Deep Belief Networks (DBN), and Recurrent Neural Networks (RNN) \cite{8}. As deep learning technology has advanced, more sophisticated models have been developed for detecting anomalies in cybersecurity. For instance, Singh \textit{et al.} \cite{161} illustrate the benefits of AD in wide-area protection schemes (WAPS) by using a deep learning-based cyber-physical AD system (CPADS) to detect and mitigate data integrity and communication failure attacks in centralized Remedial Action Schemes (CRAS). Similarly, Nagarajan \textit{et al.} \cite{162} highlights the effectiveness of AD in enhancing the security of Cyber-Physical Systems (CPSs) by accurately identifying anomalous behaviors, thereby addressing the growing challenges posed by sophisticated cyber-attacks and the increasing volume of data.
\subsection{Applications in Visual data}
AD in visual data, encompassing images and videos, plays a vital role in numerous industries where visual inspection is critical. Applications range from detecting defects in manufacturing processes to identifying medical abnormalities in imaging, monitoring public safety through surveillance systems, and ensuring quality control in production lines. By leveraging advanced deep learning techniques, AD methods can automatically identify and analyze irregularities with high precision, reducing reliance on manual inspection and improving efficiency. In this section, we explore key applications of visual data-based AD, highlighting its transformative impact across various domains.
\subsubsection{Medical Imaging} 
AD in medical imaging is indispensable across numerous medical specialties, playing a crucial role in the early detection and diagnosis of diseases. In radiology, it is employed to identify anomalies in X-rays \cite{164}, brain imaging \cite{163}, and CT scans \cite{165}, thereby aiding in the accurate diagnosis of various conditions. However, as \cite{166} highlights, anomalies in medical images often closely resemble normal tissue, posing a significant challenge to detection due to their subtle differences. This similarity requires the use of sophisticated techniques to effectively distinguish between normal and anomalous data. For example, Draelos \textit{et al.} \cite{167} demonstrate the power of machine learning in radiology, significantly enhancing the classification performance for multiple abnormalities in chest CT volumes, achieving an AUROC greater than 0.90 for 18 different abnormalities. Additionally, Shvetsova \textit{et al.} \cite{130} showcase a novel method for AD in medical images, which dramatically improves the detection of subtle abnormalities in complex, high-resolution images, such as chest X-rays and pathology slides—scenarios where traditional models often fail. Furthermore, Zhao \textit{et al.} \cite{147} introduce the SALAD framework, which enhances AD in medical images by utilizing self-supervised and translation-consistent features from normal data. This approach is particularly effective in situations where labeled anomalous images are scarce, thereby improving detection accuracy in challenging medical imaging tasks.
\subsubsection{Video Monitoring} 
Video AD (VAD) has become increasingly crucial with the rise of large-scale multimedia data analysis, particularly in the processing of video data \cite{168}. VAD focuses on identifying unusual patterns or behaviors in video footage that deviate from the norm, making it a vital tool in several domains. In security and surveillance, VAD is used to monitor public spaces, buildings, and secure areas, enabling the detection of suspicious activities, unauthorized access, and unusual crowd behaviors, thereby enhancing public safety \cite{169}. In the realm of traffic monitoring, VAD facilitates the real-time identification of accidents and irregular traffic patterns, allowing for prompt response and management \cite{145}. Additionally, VAD is applied in behavioral analysis to detect abnormal behaviors in various environments, such as schools, workplaces, and public transportation systems, contributing to the maintenance of safety and order. For example, Chen \textit{et al.} \cite{105} propose a bidirectional prediction framework specifically designed for AD in surveillance videos. This innovative approach employs forward and backward prediction subnetworks to predict the same target frame, constructing a loss function based on the real target frame and its bidirectional predictions. Experimental results demonstrate that this model outperforms existing approaches on various surveillance video datasets, including those featuring pedestrians and street scenes, showcasing its superior performance in accurately detecting anomalies in real-world surveillance scenarios.

\section{Deep learning methods for Anomaly Detection} \label{sec:deep}
The application of deep learning to AD has revolutionized the way we identify irregularities in both time-based and non-time-based datasets \cite{189}. Traditional methods, such as statistical analysis and clustering, have been commonly used to detect anomalies. However, these methods often struggle with high-dimensional data, complex relationships, and capturing intricate patterns. Deep learning models, with their ability to learn hierarchical representations and detect subtle anomalies, have emerged as powerful tools to overcome these limitations.

\begin{figure}[!t]
    \centering
    \includegraphics[width=3in]{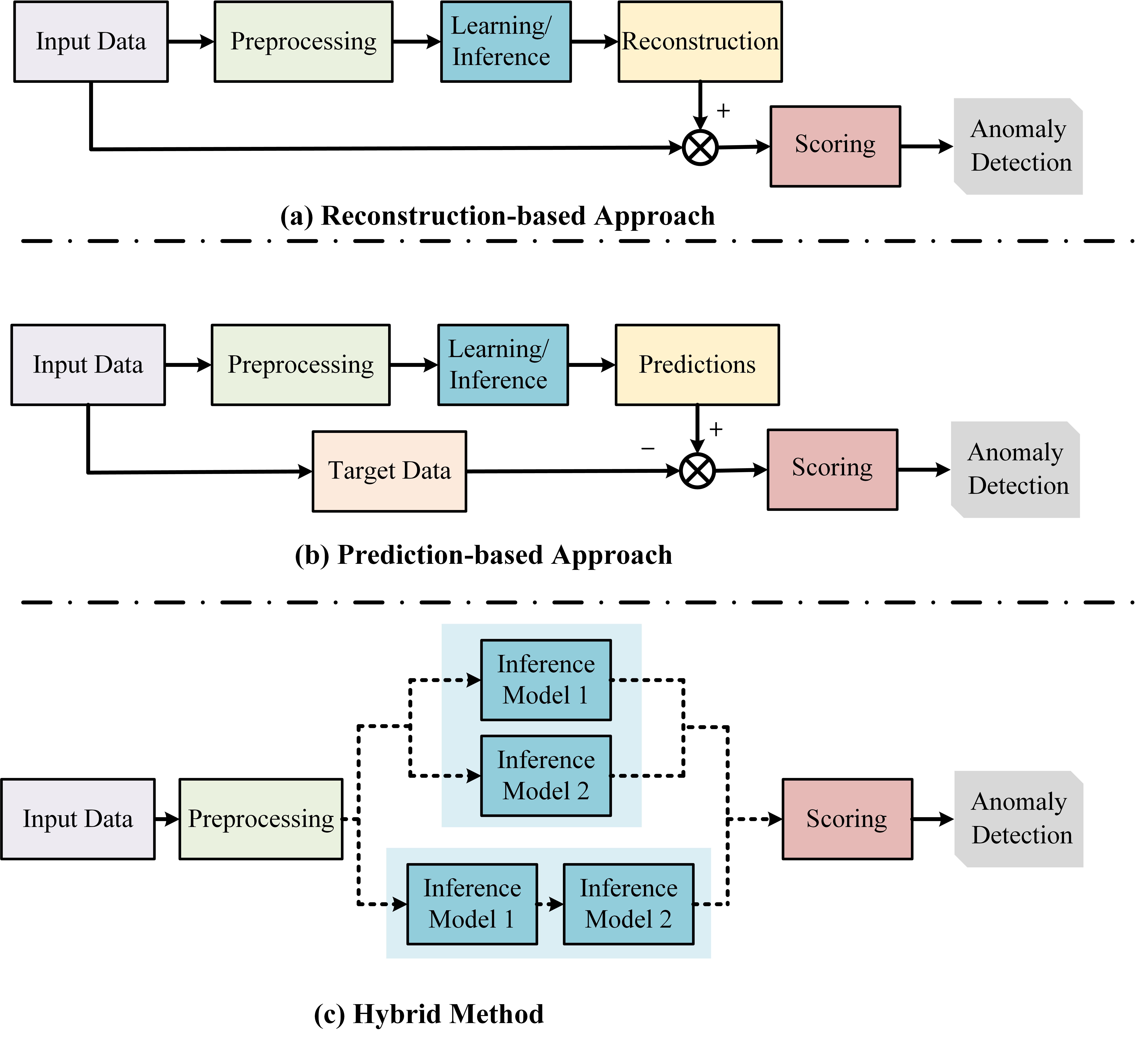}
	\caption{Three types of anomaly detection: (a) Reconstruction-based approache, (b) Prediction-based approache, (c) Hybrid method.}
	\label{threetype_for_TSAD}
\end{figure}

As shown in Fig.\ref{threetype_for_TSAD}, this section introduces three major deep learning approaches applied to AD: reconstruction-based methods, prediction-based methods, and hybrid approaches. Each approach leverages the strengths of deep learning in distinct ways to improve AD accuracy, particularly in scenarios where data patterns are complex, unstructured, or temporal.

\begin{table*}[!t]
\caption{Comparison of GANs, VAEs, and Diffusion Models in Anomaly Detection\label{tab:GVD}}
\centering
\scriptsize
\begin{tabular}{|>{\centering\arraybackslash}m{2cm}|>{\raggedright\arraybackslash}m{7cm}|>{\raggedright\arraybackslash}m{7cm}|}
\hline
\textbf{Model} & \textbf{Strengths} & \textbf{Weaknesses} \\
\hline
\textbf{GANs} & 
\textbf{•} Capable of generating high-fidelity, realistic samples. \newline
\textbf{•} Learns complex data distributions using adversarial loss. \newline
\textbf{•} Useful in AD by distinguishing real vs. generated data. & 
\textbf{•} Prone to mode collapse, leading to low sample diversity. \newline
\textbf{•} Hard to train with difficult-to-interpret losses. \newline
\textbf{•} Training is unstable and hard to converge. \\
\hline
\textbf{VAEs} & 
\textbf{•} Easy to train with one tractable likelihood loss. \newline
\textbf{•} Provides high sample diversity by covering all data modes. \newline
\textbf{•} Latent space representation is useful for AD tasks. & 
\textbf{•} Produces low-fidelity, often blurry samples. \newline
\textbf{•} Pixel-based loss leads to sample ambiguity and blurriness. \\
\hline
\textbf{Diffusion Models} & 
\textbf{•} Generates high-fidelity samples with gradual refinement. \newline
\textbf{•} High sample diversity due to likelihood maximization. \newline
\textbf{•} Intermediate noisy images serve as useful latent codes for AD. & 
\textbf{•} Slow sample generation due to the multi-step denoising process. \newline
\textbf{•} Computationally intensive, requiring many steps for both forward and reverse diffusion. \\
\hline
\end{tabular}
\end{table*}

\subsection{Deep learning methods for Anomaly Detection based on Reconstruction}
Reconstruction-based approaches operate by training a model to learn the underlying distribution of normal data \cite{190}. Once trained, the model attempts to reconstruct incoming data. The reconstruction error, which is the difference between the original data and its reconstruction, is then used as an indicator of anomaly. A high reconstruction error suggests that the data is anomalous, as it deviates from the learned normal patterns. Deep learning-based reconstructive models have become prominent due to their ability to capture complex patterns in high-dimensional data. In recent years, most reconstruction-based AD models have been developed using techniques such as GAN, AE, and diffusion models. These models each have unique strengths and weaknesses, as summarized in Table \ref{tab:GVD}. This table consolidates insights from multiple studies, including \cite{cao2024survey}, \cite{goodfellow2020generative}, \cite{yang2023diffusion}, and \cite{bond2021deep}, which have analyzed the advantages and limitations of GANs, VAEs, and Diffusion Models in AD. In this section, we introduce these three types of models in the context of AD and discuss their various variants.
\subsubsection{GAN-based Anomaly Detection}

\begin{figure}[!t]
\centering
\includegraphics[width=3in]{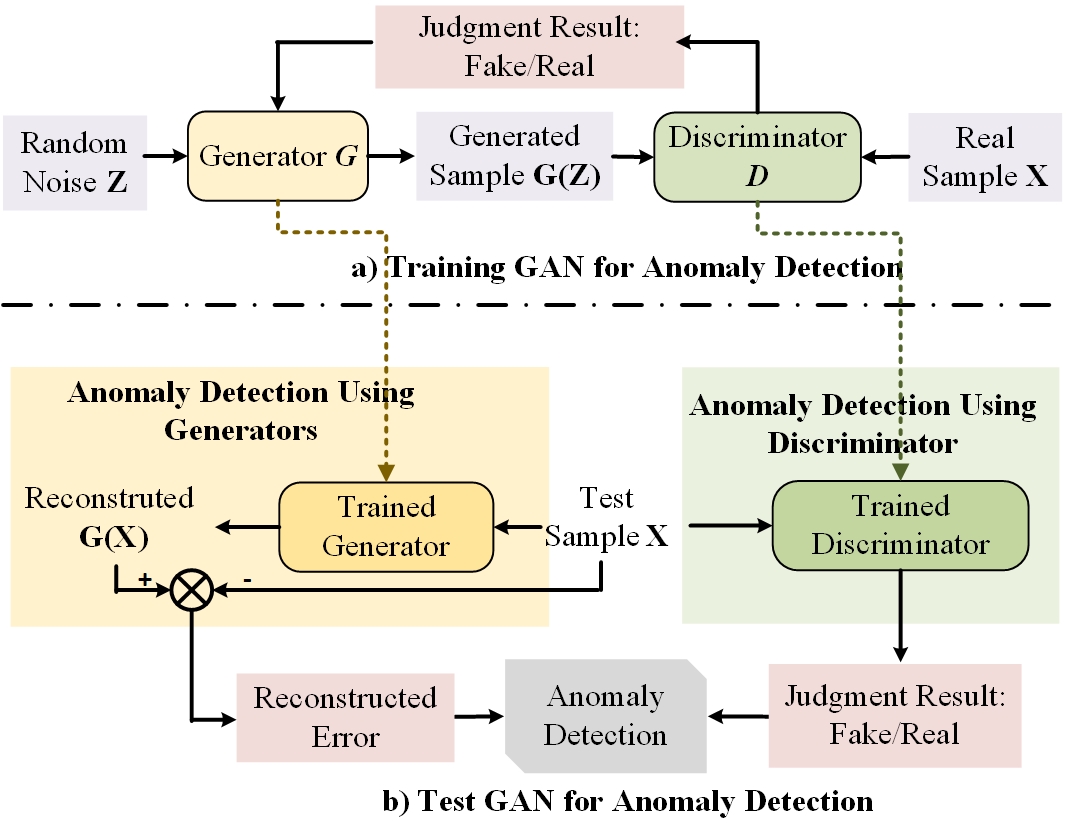}
\caption{Structural Frameworks for GAN Anomaly Detection.}
\label{GAN}
\end{figure}

GANs are powerful tools for generating synthetic data that resembles a given training dataset \cite{152}. As shown in the upper part of Fig.\ref{GAN}, GANs consist of two main components: a generator and a discriminator, both of which are neural networks. Because of this structure, GAN models are highly flexible, allowing for different networks to be chosen as the generator and discriminator based on the specific task. This flexibility makes GANs a versatile framework for a wide range of applications. The generator \(G\) takes a random noise vector \(z\) (usually sampled from a Gaussian distribution) as input and generates synthetic data \(G(z)\). The discriminator \(D\) receives a data sample (either from the real dataset or from the generator) as input and outputs a probability \(D(x)\), representing the likelihood that the input is real (i.e., from the actual dataset) rather than fake (i.e., generated by the generator). The generator and discriminator are trained simultaneously through a process where the generator tries to produce data that can fool the discriminator, and the discriminator tries to improve its ability to distinguish between real and fake data. Table \ref{tab:GAN} provides a comprehensive summary of recent GAN-based AD models, categorizing them based on their techniques, approaches, strengths, and weaknesses. This table highlights how different GAN variants are tailored for specific AD tasks, along with the types of data they are applied to and their publication years.

The training process of GANs can be described as a minimax game with the following objective function:

\begin{align}
\min_G \max_D V(D, G) &= \mathbb{E}_{x \sim p_{data}(x)}[\log D(x)] \notag\\
&\quad + \mathbb{E}_{z \sim p_z(z)}[\log(1 - D(G(z)))].
\end{align}

In this function, \(p_{data}(x)\) represents the distribution of the real data, \(p_z(z)\) represents the distribution of the noise vector \(z\), \(G(z)\) is the data generated by the generator, and \(D(x)\) is the probability that \(x\) is real. The generator \(G\) aims to minimize this objective, while the discriminator \(D\) aims to maximize it. The discriminator updates its weights to maximize the probability of correctly classifying real and generated data, while the generator updates its weights to minimize the discriminator's ability to distinguish between real and fake data.


In the context of AD, GANs play crucial roles in both representation learning and data augmentation, each serving distinct purposes within deep Learning \cite{33}. In representation learning, the primary objective of GANs is to learn and model the underlying distribution of the data, enabling the generation of synthetic data that closely resembles real data. This process involves a generator that creates fake data from random noise and a discriminator that distinguishes between real and fake data. Through iterative training, the generator improves its ability to produce realistic data, which is particularly useful in tasks like AD. For example, in \cite{39}, GANs are used for representation learning by generating fake data that matches the distribution of normal data. This generated data is then used to train a VAE to detect anomalies through reconstruction errors. Similarly, in \cite{141}, a fault-attention generative probabilistic adversarial autoencoder (FGPAA) is proposed, combining GANs and autoencoders for AD by learning the low-dimensional manifold of healthy state data. The GAN component aids in feature representation learning, reducing signal information loss and enhancing the model’s ability to detect anomalies through distribution probability and reconstruction error. 

There are two main structures to using GANs for AD, as shown in Fig.\ref{GAN}. The first approach is based on the generator, as depicted in the lower part of Fig.\ref{GAN}, highlighted by the yellow box. The basic idea is to train the GAN on normal data and then use the reconstruction error to identify anomalies. During the training phase, the GAN is trained exclusively on normal data, allowing the generator to learn to produce data that closely mimics the normal data distribution. During the detection phase, a test data point \( x \) is fed into the generator to obtain the reconstructed data \( G(x) \). The reconstruction error, typically measured as the difference between the original data point \( x \) and the reconstructed data \( G(x) \), is then used to detect anomalies. This can be quantified using metrics such as mean squared error (MSE). If the reconstruction error exceeds a predefined threshold, the data point is classified as an anomaly. The intuition behind this approach is that the generator, trained solely on normal data, will struggle to accurately reconstruct anomalous data, resulting in a high reconstruction error. 

The mathematical representation for AD using GANs involves computing the reconstruction error \( E(x) \) as follows:

\begin{equation}
E(x) = \| x - G(x) \|^2,
\end{equation}
where \(\| \cdot \|^2\) denotes the squared Euclidean distance. A threshold \(\tau\) is set, and if \( E(x) > \tau \), the data point \( x \) is considered an anomaly. For example, Dong \textit{et al.} \cite{158} propose a semi-supervised approach for video AD using a dual discriminator-based GAN structure, focusing on representation learning. In this approach, the generator predicts future frames for normal events, and anomalies are detected by evaluating the quality of these predictions. Similarly, Guo \textit{et al.} \cite{35} introduce RegraphGAN, a graph generative adversarial network specifically designed for dynamic graph AD. RegraphGAN utilizes GAN-based representation learning to encode complex spatiotemporal relationships in graph data, allowing it to better capture anomalies. By leveraging encoders to project input samples into a latent space and integrating GANs to enhance both training stability and efficiency, RegraphGAN significantly improves AD performance over existing methods.

The second approach leverages the discriminator highlighted by the green box in Fig.\ref{GAN}. A well-trained discriminator has the ability to differentiate between real (normal) and fake (anomalous) samples. During the detection phase, test samples are directly input to the discriminator, which evaluates the likelihood that a given sample is real. If the discriminator assigns a low probability to a sample, suggesting that it is likely fake or anomalous, the sample is flagged as an anomaly. This method relies on the discriminator's capacity to recognize deviations from the normal data distribution it learned during training. For instance, Liu \textit{et al.}\cite{36} propose a GAN framework that uses multiple generators to produce potential outliers, which are then distinguished from normal data by a discriminator to detect anomalies. The discriminator's output score is used to evaluate the anomaly degree of input data, providing a comprehensive reference distribution and preventing mode collapse.

Additionally, GANs are highly effective in data augmentation, helping to mitigate the scarcity of anomaly samples, which often results in data imbalance and poor generalization \cite{184}. When anomaly samples are unevenly distributed or lacking in diversity, models struggle to learn rare anomalies and can overfit to the training set, reducing their accuracy on unseen data. Traditional data augmentation techniques—such as scaling, rotation, random cropping, translation, flipping, and copy-paste—attempt to mitigate these issues. However, simple linear transformations fail to capture new distributions and features of unknown anomalies, such as random changes in shape or texture. This is where GANs provide a significant advantage. By generating synthetic anomaly data that mimics the distribution of real-world anomalies, GANs enable models to learn a more diverse set of anomaly features. This not only addresses the imbalance problem but also improves the model’s generalization capabilities, as it learns to detect anomalies based on a broader range of characteristics beyond those present in the original training dataset. Miao \textit{et al.} \cite{78} introduce an unsupervised AD framework that uses data augmentation through contrastive learning and GANs to mitigate overfitting. By employing a geometric distribution mask, it enhances data diversity and generates synthetic anomaly samples, addressing the scarcity of anomaly data. In \cite{185}, Anomaly-GAN addresses data augmentation by using a mask pool, anomaly-aware loss, and local-global discriminators to generate high-quality, realistic synthetic anomalies with diverse shapes, angles, spatial locations, and quantities in a controllable manner. Li \textit{et al.} \cite{186} propose augmented time regularized generative adversarial network that combines an augmented filter layer and a novel temporal distance metric to generate high-quality and diverse artificial data, addressing the limitations of existing GAN approaches in handling limited training data and temporal order. 

\begin{table*}[!t]
\caption{GAN-based Models in Anomaly Detection\label{tab:GAN}}
\centering
\scriptsize
\begin{tabular}{|>{\centering\arraybackslash}m{0.5cm}|>{\centering\arraybackslash}m{2cm}|>{\centering\arraybackslash}m{2cm}|>{\raggedright\arraybackslash}m{4cm}|>{\raggedright\arraybackslash}m{4cm}|>
{\centering\arraybackslash}m{2cm}|>{\centering\arraybackslash}m{0.8cm}|}
\hline
\textbf{Paper} & \textbf{Technique} & \textbf{Approach Type} & \textbf{Strength} & \textbf{Weakness} & \textbf{Data Type}& \textbf{Year} \\
\hline
\cite{36} & GAN & Reconstruction & Does not depend on assumptions about the normal data and requires less computing resources. & The method involves the selection of multiple hyperparameters, making the tuning process challenging and potentially time-consuming. & Structured data & 2020 \\
\hline
\cite{181} &GAN+CNN & Prediction & The NM-GAN model enhances both the generalization and discrimination abilities through noise-modulated adversarial learning, resulting in improved accuracy and stability for video AD. & The model struggles to fully capture complex temporal patterns like staying, wandering, and running, and lacks adaptive modulation of generalization and discrimination abilities, leaving room for improvement in spatiotemporal feature learning. & Video data & 2021\\
\hline

\cite{186} & GAN & Reconstruction & Is capable of generating more effective artificial samples for training supervised learning models, thereby addressing the issue of data imbalance. & Its performance is inferior to the baseline algorithms when the balanced ratio is 0.125. & Image data & 2021 \\
\hline
\cite{82} & GAN+LSTM & Prediction & The TMANomaly framework excels in capturing complex multivariate correlations in industrial time series data, enhancing AD accuracy through mutual adversarial training. & The paper lacks discussion on TMANomaly's generalization to other datasets, the potential limitations of using GRA for feature selection, and the computational efficiency or scalability, which are critical for real-time industrial systems. & Multivariate time series data & 2022\\
\hline

\cite{83} & GAN+LSTM & Prediction & FGANomaly method effectively filters anomalous samples before training, improving AD accuracy and robustness by precisely capturing normal data distribution and dynamically adjusting generator focus.& The method lacks effective fusion of information across different dimensions in multivariate time series, which limits its ability to fully capture complex correlations. & Multivariate time series data & 2022\\
\hline
\cite{185} & GAN & Reconstruction & Improves the quality of the generated anomaly images and generates anomalies with different shapes, rotation angles, spatial locations, and numbers in a controllable manner. & The images generated are not very sensitive to the change of light. & Image data & 2023 \\
\hline
\cite{35} & GAN & Reconstruction & Improves training efficiency and stability in dynamic graph AD while avoiding the expensive optimization process typical of traditional graph generative adversarial networks. & The detection accuracy on the UCI Message dataset is lower than that of TADDY. & Dynamic graph data & 2023 \\
\hline
\cite{78} & GAN+Transformer & Reconstruction & It can effectively detect anomalies in long sequences, mitigates overfitting, and incorporates contrastive loss into the discriminator to fine-tune the GAN, ensuring strong generalization ability. & It may struggle with irregularly sampled data or datasets with many missing values, requires careful tuning of several hyperparameters, and demands significant computational resources, posing challenges for real-time processing on limited-capacity devices. & Multivariate time series data & 2024 \\
\hline

\end{tabular}
\end{table*}

\subsubsection{AE-based Anomaly Detection}

In recent years, the limitations of traditional AE models in handling complex and noisy data have become more apparent, leading to the development of enhanced methods to improve their performance in AD tasks. For example, Fan \textit{et al.} \cite{131} introduce a new framework by incorporating \( \ell_{2,1} \)-norm into the AE, and experiments have demonstrated that this framework can significantly improve ADn accuracy by increasing the model's robustness to noise and outliers during training. Wang \textit{et al.} \cite{43} demonstrate that introducing an adaptive-weighted loss function can effectively suppress anomaly reconstruction, thereby improving the accuracy of AD. Liu \textit{et al.} \cite{37} introduce a multi-scale convolutional AE architecture, where multiple stacked convolutional encoder-decoder layers act as background learners to robustly eliminate anomalies of varying sizes during background reconstruction. Additionally, Lin \textit{et al.} \cite{69} introduce a soft calibration strategy combined with AE to address the issue of data contamination in AD.

\begin{figure}[!t]
\centering
\includegraphics[width=3in]{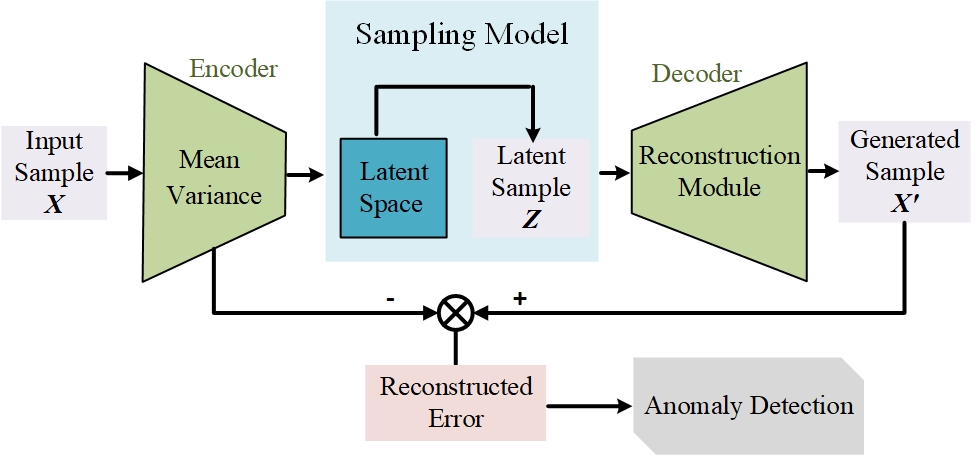}
\caption{Structural Frameworks for VAE Anomaly Detection.}
\label{VAE}
\end{figure}

VAEs are another generative model widely used in AD tasks. Like GANs, VAEs aim to learn the distribution of normal data to identify anomalies. However, unlike GANs, which rely on adversarial training between a generator and a discriminator, VAEs use an encoder-decoder architecture. Fig.\ref{VAE} illustrates the structure of AD based on VAE. The goal of a VAE is to map the input data into a latent space through the encoder and model the data distribution probabilistically within this space. This approach allows the VAE to generate new data that closely resembles the true data distribution, and anomalies can be detected by evaluating the reconstruction error.

 The internal structure of a VAE is similar to that of a traditional AE but with some key differences. First, the encoder in a VAE not only compresses the input data into a lower-dimensional latent space but also learns a probabilistic distribution, typically parameterized by a mean \( \mu \) and a variance \( \sigma^2 \) as shown in Fig.\ref{VAE}. This enables the VAE to generate more meaningful latent variables \( z \), enhancing the diversity and robustness of the generated data.
A critical component introduced in VAEs is the Kullback-Leibler (KL) divergence, which measures the difference between the latent distribution generated by the encoder and a predefined prior distribution (usually a standard normal distribution). Unlike traditional AEs, which focus solely on minimizing the reconstruction error, VAEs are trained by minimizing a combination of the reconstruction error and the KL divergence:
\begin{equation}
\mathcal{L}_{\text{VAE}} = \mathbb{E}_{q(z|x)}[\log p(x|z)] - D_{\text{KL}}(q(z|x) \| p(z)).
\end{equation}

This difference makes VAEs more powerful in AD because they not only consider the quality of the data reconstruction but also enforce a structured latent space through the KL divergence. By doing so, KL divergence helps to regularize the latent space, ensuring that the encoded representations are smoothly distributed and centered around the prior distribution. This regularization reduces overfitting, promotes better generalization, and makes it easier to distinguish between normal and anomalous data, especially in complex and high-dimensional datasets. Table \ref{tab:AE} provides a comprehensive summary of the latest advancements in VAE-based AD models, showcasing innovative enhancements that address various challenges such as noise robustness, semantic feature learning, and anomaly reconstruction. Huang \textit{et al.} \cite{124} enhance VAE-based AD by incorporating an Autoencoding Transformation into the model, which ensures that the training phase effectively captures high-level visual semantic features of normal images, thereby increasing the anomaly score gap between normal and anomalous samples. Similarly, Yin \textit{et al.} \cite{57} utilize Convolutional Neural Network (CNN) and VAE with a two-stage sliding window approach in data preprocessing to learn better representations for AD tasks. Zhang Yin \textit{et al.} \cite{zhang2022grelen} propose the Graph Relational Learning Network (GReLeN), which integrates a VAE structure with graph dependency learning for AD in multivariate time series through reconstruction.
Zhou \textit{et al.} \cite{42} propose a variational long short-term memory (VLSTM) model for high-dimensional AD in imbalanced datasets, combining a compression network for efficient data representation with an estimation network for accurate classification of network traffic data. The VLSTM model balances data compression and feature retention using core LSTM and variational modules. 

In recent years, many advancements in AD models inspired by VAEs have focused on Adversarial Autoencoders (AAEs) \cite{38}. Unlike traditional VAEs, which use KL divergence to match the latent space distribution to a prior, AAEs achieve this through the use of GANs. Specifically, AAEs employ a GAN's discriminator to evaluate the latent variable distribution produced by the encoder and use adversarial training to align it with the desired prior distribution, providing more flexible control over the quality of the generated data. Wu \textit{et al.} \cite{141} propose the Fault-Attention Generative Probabilistic Adversarial Autoencoder (FGPAA) for machine AD, utilizing an end-to-end AAE with double discriminators to extract relevant features and ensure accurate equipment health monitoring through a fault-attention probability distribution. Idrissi \textit{et al.} \cite{40} apply AAE and FL in the field of network intrusion detection, effectively ensuring AD performance while safeguarding client privacy. Experimental results demonstrate that the proposed model outperforms AE, VAE, and AAE on various network traffic datasets, achieving high performance across different metrics. Su \textit{et al.} \cite{41} propose two contamination-immune BiGAN models, integrating elements of VAE and BiGAN to create a new AAE-based framework that effectively detects anomalies by learning the probability distribution of normal samples from contaminated datasets, significantly outperforming state-of-the-art methods in scenarios where training data is impure. Similar to the aforementioned AAE models, Du \textit{et al.} use GANs to purify the original dataset, generating synthetic ``normal'' data to improve outlier detection accuracy. Continuing the advancements in AAE-based models, Yu \textit{et al.} \cite{62} introduce an Adversarial Contrastive Autoencoder (ACAE) for Multivariate Time Series (MTS) AD, which enhances feature representation through adversarial training and contrastive learning, demonstrating superior performance across multiple real-world datasets, further extending the application of AAE-based methods in robust AD.

\begin{table*}[!t]
\caption{Autoencoder-based Models in Anomaly Detection\label{tab:AE}}
\centering
\scriptsize
\begin{tabular}{|>{\centering\arraybackslash}m{0.5cm}|>{\centering\arraybackslash}m{2cm}|>{\centering\arraybackslash}m{1.8cm}|>{\raggedright\arraybackslash}m{4cm}|>{\raggedright\arraybackslash}m{4cm}|>{\centering\arraybackslash}m{2cm}|>{\centering\arraybackslash}m{0.5cm}|}
\hline
\textbf{Paper} & \textbf{Technique} & \textbf{Approach Type} & \textbf{Strength} & \textbf{Weakness} & \textbf{Data Type} & \textbf{Year} \\
\hline
\cite{42} & VAE-based (VAE+LSTM) & Reconstruction & Effectively addresses imbalanced and high-dimensional challenges in industrial big data. & Falls short in achieving the highest AUC and F1 scores compared to other methods. & Industrial big data & 2020 \\
\hline
\cite{141} & AAE-based & Reconstruction & FGPAA reduces information loss during feature extraction and constructs fault attention anomaly indicators using low-dimensional feature probability and reconstruction error. & Runtime is approximately five times longer than SOM. & Rotating machine fault simulator data & 2020 \\
\hline
\cite{43} & AE-based (AE+CNN) & Reconstruction & The Auto-AD method enables fully autonomous hyperspectral AD, automatically separating anomalies based on reconstruction errors without the need for manual tuning or additional processing. & Lower AUC score compared to the GRX method on the Honghu dataset. & Hyperspectral data & 2021 \\
\hline
\cite{37} & AE-based (AE+CNN) & Reconstruction & MSNet offers an effective solution to handle multiscale anomaly shapes, providing greater flexibility without the need for threshold fine-tuning. & Multiple convolutional encoder-decoder layers and enhanced training increase computational cost and training time. & Hyperspectral data & 2021 \\
\hline
\cite{124} & VAE-based (VAE+Transformer) & Reconstruction & SSR-AE leverages self-supervised learning to enhance normal data reconstruction and hinder abnormal data, optimizing mutual information for effective transformation and image reconstruction. & Struggles with transformations, heavily relying on their effectiveness for AD. & Image data & 2021 \\
\hline
\cite{131} & AE-based & Reconstruction & Maintains geometric structure and local spatial coherence of hyperspectral images (HSI), reducing search space and execution time per pixel. & High execution time for constructing the SuperGraph matrix with large datasets. & Hyperspectral data & 2021 \\
\hline
\cite{40} & AAE-based (AAE+Federated learning) & Reconstruction & Fed-ANIDS demonstrates strong generalization, outperforms GAN-based models, and ensures privacy protection through federated learning. & Computational overhead due to the federated learning framework, increasing training complexity and latency. & Cybersecurity data & 2023 \\
\hline
\cite{69} & AE-based & Reconstruction & Applicable for time series AD under data contamination. & Assumes normal samples follow a Gaussian distribution, limiting applicability, and has higher computational complexity. & Time series data & 2024\\
\hline
\cite{41} & AAE-based & Reconstruction & Learns the probability distribution of normal samples from contaminated datasets, achieving convergence and outperforming baseline models. & Relies on the assumption that the contamination ratio is known, which may not always be accurate in practice. & Medical image data & 2024 \\
\hline

\cite{39} & AAE-based & Reconstruction & Generates a clean dataset from contaminated data for AD, with linear scalability for larger datasets. & Struggles with detection accuracy in datasets with multiple distribution patterns. & Tabular data & 2024 \\
\hline
\cite{62} & AAE-based & Reconstruction & Excels in learning high-level semantic features and capturing normal patterns of MTS with contrastive learning constraints, ensuring stability across parameter settings. & Performance on all metrics for SMAP and PSM datasets is lower than baseline methods. & Multivariate time series data & 2024 \\
\hline
\end{tabular}
\end{table*}

\subsubsection{Diffusion model-Based for Anomaly Detection}
Diffusion models are a type of generative model that operate through two key phases: a fixed forward diffusion process and a learnable reverse diffusion process \cite{45}. Mathematically, the forward process involves progressively adding Gaussian noise to the data $x_0$, transforming it into pure noise $x_T$ over $T$ steps. This process can be described as:
\begin{equation}   
q(x_t | x_{t-1}) = \mathcal{N}(x_t; \sqrt{1 - \beta_t} x_{t-1}, \beta_t I)
,
\end{equation}
where $q(x_t | x_{t-1})$ is the conditional probability distribution of $x_t$ given $x_{t-1}$, $\beta_t$ is the noise variance at step $t$, and $x_t$ represents the noisy data at step $t$. As $t$ increases, the data becomes more corrupted by noise until it reaches a state of pure Gaussian noise at step $T$.

The reverse process learns to gradually denoise the data, removing the added noise step by step. The model learns a parameterized distribution $p_\theta(x_{t-1} | x_t)$ to reverse the noise addition process, reconstructing the original data from the noisy data. This reverse process is trained to minimize the variational bound on the data likelihood, expressed as:
\begin{equation}
L = \mathbb{E}_q \left[ D_{KL}(q(x_{t-1} | x_t, x_0) || p_\theta(x_{t-1} | x_t)) \right].
\end{equation}
By progressively removing noise, diffusion models generate high-fidelity samples, first capturing coarse structures and then refining details in each step. In the context of AD, diffusion models are trained on normal data to learn the underlying data distribution through an iterative noise-removal process. Similar to other reconstruction-based methods, anomalies can be identified by evaluating the reconstruction error, where a higher error indicates that the data deviates from the learned normal patterns.

Diffusion models stand out from GANs and VAEs in several key ways. They avoid common issues such as mode collapse in GANs, where only a subset of the data distribution is captured, leading to reduced diversity. Diffusion models also overcome the blurriness associated with VAEs, which often results from pixel-based loss and a smaller latent space. By iteratively denoising data, diffusion models maintain both high fidelity and diversity in their outputs.

While diffusion models are slower in generating samples due to their iterative nature, their ability to accurately reconstruct data and cover the full range of the training dataset makes them particularly well-suited for AD \cite{94}. In AD, where precision is critical, diffusion models excel by generating detailed and high-quality samples, enabling them to identify subtle deviations from normal patterns with greater accuracy than other generative models. Several works have leveraged the advantages of diffusion models in ADn. For example, Zhang \textit{et al.} \cite{46} utilize the high-quality and diverse image generation capabilities of diffusion models to enhance reconstruction quality in DiffAD, addressing the limitations of traditional methods by introducing noisy condition embedding and interpolated channels. Similarly, Li \textit{et al.} \cite{47} apply a diffusion model to reconstruct normal data distributions and integrate an auxiliary learning module with pretext tasks to better distinguish between normal and abnormal data. Expanding on these ideas, Zeng \textit{et al.} \cite{49} improve denoising diffusion probabilistic models (DDPMs) for radio AD by incorporating an AE to learn the distribution of normal signals and their power spectral density (PSD), using reconstruction error to identify anomalies. Li \textit{et al.} \cite{52} present a Controlled Graph Neural Network (ConGNN) approach based on DDPMs to address the challenge of limited labeled data. Li \textit{et al.} \cite{50} further explore diffusion models in vehicle trajectory AD, employing decoupled Transformer-based encoders to capture temporal dependencies and spatial interactions among vehicles, significantly improving AUC and F1 scores on real-world and synthetic datasets. Similarly, Pei \textit{et al.} \cite{ref:imputation} establish the two-stage diffusion model (TSDM) to mitigate the influences of anomalies in smart grids, where the first stage is a diffusion-based AD component. In multi-class AD, He \textit{et al.} \cite{51} propose DiAD, a framework that enhances reconstruction accuracy through a combination of a semantic-guided network, spatial-aware feature fusion, and a pre-trained feature extractor to generate anomaly maps.

\begin{table*}[!t]
\caption{Diffusion-based Models in Anomaly Detection\label{tab:AEt}}
\centering
\scriptsize
\begin{tabular}{|>{\centering\arraybackslash}m{0.5cm}|>{\centering\arraybackslash}m{2.2cm}|>{\centering\arraybackslash}m{1.5cm}|>{\raggedright\arraybackslash}m{4cm}|>{\raggedright\arraybackslash}m{4cm}|>{\centering\arraybackslash}m{1.5cm}|>{\centering\arraybackslash}m{0.5cm}|}
\hline
\textbf{Paper} & \textbf{Technique} & \textbf{Approach Type} & \textbf{Strength} & \textbf{Weakness} & \textbf{Data Type} & \textbf{Year} \\
\hline
\cite{46} & Diffusion & Reconstruction & The latent diffusion model (LDM) used in this method achieves state-of-the-art performance in surface AD by generating high-quality, semantically correct reconstructions, effectively avoiding overfitting to anomalies. & It less suitable for real-time applications or environments with limited computational resources. & Image data & 2023\\
\hline
\cite{49} & Diffusion+VAE & Reconstruction &The AE-DDPMs algorithm effectively improves stability and reduces computational costs in radio AD, outperforming GAN-based methods in complex electromagnetic environments. & The anomalies in the experimental data are artificially generated, rather than originating from real-world conditions, which may limit the model's applicability to genuine, real-world scenarios. & radio signal data & 2023\\
\hline
\cite{52} & Diffusion+GNN & Prediction & ConGNN effectively addresses the issue of limited labeled data by generating augmented graph data using a graph-specific diffusion model. & The reliance on graph-specific augmentation might not generalize well to other types of data, potentially limiting its applicability beyond graph-based AD. & Image data & 2023\\
\hline
\cite{47}  & Diffusion+VAE & Hybrid & SDAD effectively enhances AD by combining self-supervised learning for discriminative data representation with denoising diffusion. & The generation of pseudo anomalies relies solely on standard Gaussian sampling, which may not fully capture the complexity of real anomalies, limiting the model's ability to accurately simulate genuine abnormal data. & Structure data & 2024\\
\hline
\cite{50} & Diffusion+Transformer & Hybrid & DiffTAD effectively models temporal dependencies and spatial interactions in vehicle trajectories through diffusion models, significantly improving AD accuracy and robustness to noise. & The anomalies are primarily evaluated on synthetic datasets, which may not fully reflect the complexity and diversity of real-world trajectory data. & Vehicle trajectory data & 2024\\
\hline

\end{tabular}
\end{table*}

\subsection{Deep learning methods for Anomaly Detection based on Prediction}
Prediction-based AD methods operate by forecasting future values or estimating missing attributes and comparing these predictions to the actual observed values. When significant deviations occur, it indicates potential anomalies, as the data deviates from the learned normal patterns. These methods are versatile and can be applied across various data types, leveraging relationships between variables or temporal correlations to detect anomalies. Prediction-based methods excel in scenarios where capturing patterns and trends is essential. By learning underlying structures in the data, whether based on time dependencies or more general interactions between variables, these methods can effectively predict expected outcomes. Deviations from these expectations are flagged as anomalies. This makes prediction-based approaches highly adaptable, capable of functioning across different contexts, including various types of data. In this section, we explore three main approaches for prediction-based AD:  Recurrent Neural Networks (RNNs), attention mechanisms, and Graph Neural Networks (GNNs), all of which have demonstrated efficacy in capturing intricate patterns and relationships within data to identify anomalies. These methods allow for flexible and robust AD across various data types by learning underlying patterns, whether they are based on spatial, temporal, or graph-based relationships. By leveraging these approaches, prediction-based methods can effectively model complex interactions, providing reliable detection of unexpected behaviors or deviations from learned patterns.

\subsubsection{RNN-based Anomaly Detection}
Recurrent Neural Networks (RNNs) \cite{sherstinsky2020fundamentals} are a special type of neural network designed to process sequential data by capturing dependencies between elements in a sequence. Unlike standard neural networks, RNNs incorporate a state vector $\bm{s}_t$ in the hidden layer, allowing them to retain information from previous steps and model sequential patterns. This capability makes them effective in various applications where data has an inherent order, such as event logs, system monitoring, and structured sequences in cybersecurity or industrial processes. For an input $\bm{x}_t$ at time $t$, the update of the state value $\bm{s}_t$ and hidden layer output $\bm{h}_t$ in RNNs can be represented as 
\begin{equation}
    \begin{aligned}
\bm{s}_t &= \sigma \left (\bm{W}^x\bm{x}_t+\bm{W}^s\bm{s}_{t-1}+\bm{b}^s  \right )\\
\bm{h}_t &= \textit{\textrm{softmax}}(\bm{W}^h\bm{s}_t + \bm{b}^h),
    \end{aligned}
\end{equation}
where $\sigma(\cdot)$ is the sigmoid activation function, $\bm{W}^x$, $\bm{W}^s$ and $\bm{W}^h$ represent the network weights, and $\bm{b}$ is the network biases. By maintaining a recurrent state, RNNs can effectively capture dependencies across different steps within a sequence, making them well-suited for tasks involving ordered data.

\begin{figure*}[!t]
	\vspace{-0.2cm}
	\centerline{\includegraphics[width=0.99\textwidth]{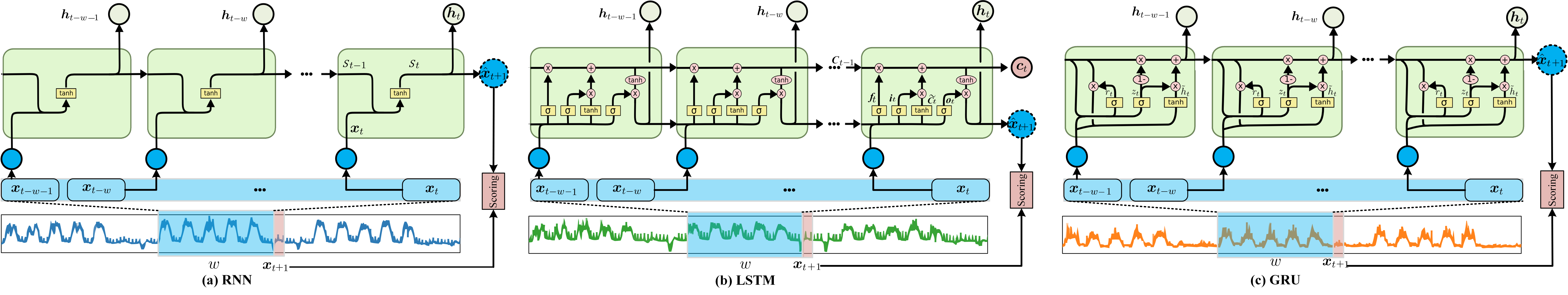}}
	\caption{RNN-based application example for time series data anomaly detection: (a) RNN-based, (b) LSTM-based, (c) GRU-based.}
	\label{RNN_for_TSAD}
	\vspace{-0.2cm}
\end{figure*}

However, RNNs face the problem of exploding or vanishing gradients when dealing with long sequences. Long Short-Term Memory networks (LSTMs) \cite{van2020review}, a specialized type of RNN, were introduced to address these issues. Specifically, LSTMs replace the hidden layer of RNNs with an LSTM block consisting of input, output, and forget gates. The inference process of LSTM at time $t$ is given by
\begin{equation}
    \begin{aligned}
\bm{f}_t &= \sigma\left (\bm{W}^{xf}\bm{x}_t + \bm{W}^{hf}\bm{h}_{t-1}+\bm{b}^f \right )\\
\bm{i}_t &= \sigma\left (\bm{W}^{xi}\bm{x}_t + \bm{W}^{hi}\bm{h}_{t-1}+\bm{b}^i \right )\\
\tilde{\bm{c}}_t &= \textit{\textrm{tanh}}\left ( \bm{W}^{x\tilde{c}}\bm{x}_t +  \bm{W}^{h\tilde{c}}\bm{h}_{t-1} +\bm{b}^{\tilde{c}} \right )\\
\bm{c}_t &= \bm{f}_t\bm{c}_{t-1}+\bm{i}_t\tilde{c}_t\\
\bm{o}_t &= \sigma \left ( \bm{W}^{xo}\bm{x}_t + \bm{W}^{ho} \bm{h}_{t-1} + \bm{b}^o \right )\\
\bm{h}_t &= \bm{o}_t \textit{\textrm{tanh}}\left ( \bm{c}_t \right ),
    \end{aligned}
\end{equation}
where $\bm{f}_t$, $\bm{i}_t$, and $\bm{o}_t$ are the forget, input and output gate weights, respectively. $\bm{c}_t$ represents the cell state of LSTM, and $\textit{\textrm{tanh}}(\cdot)$ is the hyperbolic tangent activation function. By controlling the weights of the forget, input, and output gates, LSTM determines the importance of historical time series information and the current input on the current output, thus effectively mitigating issues of gradient vanishing and allowing robust modeling of complex sequences. Reference \cite{6} provides comprehensive evidence of LSTM's effectiveness in AD across various technical systems, demonstrating its superiority in learning complex temporal behaviors and accurately identifying anomalies.

The Gated Recurrent Unit (GRU) \cite{dey2017gate} is a simplified version of LSTM that only includes an update gate and a reset gate and uses the hidden state alone to represent both short-term and long-term information. These different types of RNNs can be used in prediction-based AD tasks, with the specific detection and inference method illustrated in Fig. \ref{RNN_for_TSAD}. RNNs, LSTMs, and GRUs take time series data from $t-w$ to $t-1$ as input, and their pre-trained neural networks use these temporally ordered data to predict the single-step or multi-step future values of the univariate or multivariate time series. If the difference between the actual and predicted values is below a threshold, no anomaly is detected; if the difference exceeds the threshold, an anomaly is detected and the spatiotemporal location of the anomaly is identified.  

Current RNN-based AD primarily focuses on improving RNN algorithms tailored to AD tasks and integrating RNN with other methods for AD. The method in \cite{106} employs a pruning algorithm to reduce the number of false data points, enabling the LSTM-based AD approach to better address the challenges posed by the extremely uneven distribution of railway traffic data. LSTM combined with AE \cite{55}, VAE \cite{56}, and Singular Value Decomposition (SVD) \cite{81} has also been used to identify anomalies in Controller Area Networks (CANs) \cite{150}, electrocardiograms, and Internet monitoring data. GANs based on adversarial learning have also been integrated into the time series learning of LSTM, achieving very high performance in scenarios with few features \cite{82}, extremely imbalanced training sets, and noise interference \cite{83}. CNN is also integrated into LSTM in a serial \cite{9061946}, parallel \cite{58}, or as a foundational layer \cite{hu2023training} to better extract the spatiotemporal correlations of multidimensional time series, thereby enhancing the performance of AD. 
GRUs, compared to LSTMs, have a more streamlined architecture, resulting in lower computational complexity during training and execution of AD tasks, and they tend to perform better on certain less complex sequential data. For instance, GRUs enhance interpretability by uncovering latent correlations in multivariate time series data from industrial control system sensors \cite{74}. Similar to LSTMs, GRUs can also be combined with AEs \cite{77} or VAEs \cite{128} in an encoder-decoder architecture to mitigate the effects of noise and anomalies, thereby improving the accuracy of AD.

\subsubsection{Attention-based Anomaly Detection}
The attention mechanism was initially applied in machine translation \cite{vaswani2017attention}, with its core idea being to enable the neural network to focus on the relevant parts of the input values. While attention-based methods have shown great promise in time series AD, their applications are not limited to temporal data. These methods can effectively capture dependencies in various types of data, including spatial, spatiotemporal, and multimodal datasets. 
This flexibility broadens their use cases across different AD tasks. Compared to RNN-based approaches, they are better suited for long or complex sequences because attention can compute dependencies between all positions in the sequence simultaneously, while RNNs process sequences sequentially, step by step. 

\begin{figure*}[!t]
	\vspace{-0.2cm}
	\centerline{\includegraphics[width=0.9\textwidth]{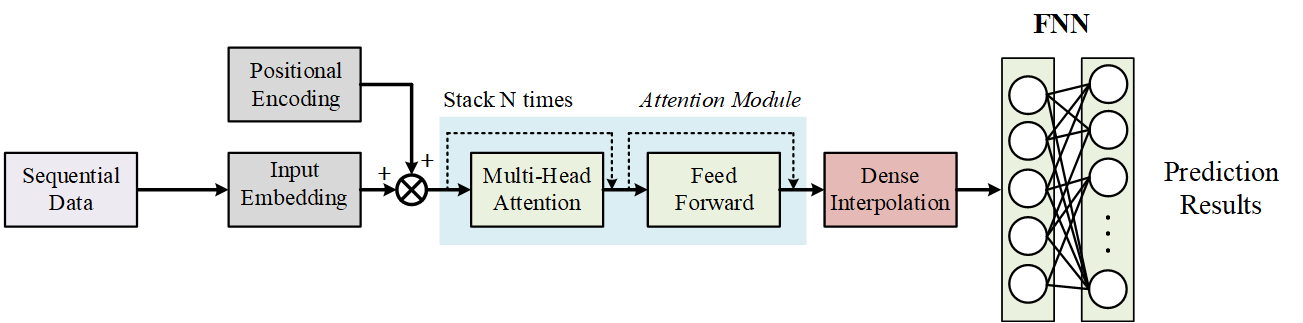}}
	\caption{Attention-based model for anomaly detection. The model first embeds sequential data using input embedding and positional encoding to preserve temporal dependencies. The multi-head attention mechanism captures long-range dependencies by processing interactions between all time steps. The feed-forward layer then refines feature representations, and a dense interpolation layer enhances anomaly-related features before passing them to a fully connected network (FNN) for final AD.}
	\label{attention_for_AD}
	\vspace{-0.2cm}
\end{figure*}

Figure \ref{attention_for_AD} illustrates a typical attention-based model for AD. Among attention-based methods, the self-attention mechanism is particularly effective in capturing global dependencies across various types of sequential data, including temporal, spatial, and spatiotemporal inputs.
For an input dataset $\bm{X}= \left [\bm{x}_1, \bm{x}_2, \cdots, \bm{x}_t \right ]$, the queries, keys, and values are defined as: $\bm{Q}= \bm{X}\bm{W}_Q$, $\bm{K}= \bm{X}\bm{W}_K$, and $\bm{V}= \bm{X}\bm{W}_V$, where $\bm{W}_Q$, $\bm{W}_K$, and $\bm{W}_V$ are trainable weight matrices.  The attention weights are then computed based on $\bm{Q}$, $\bm{K}$, and $\bm{V}$ as 
\begin{equation}
    \begin{aligned}
\alpha_{ij} = \frac{\textrm{exp}\left ( \bm{Q}_i\bm{K}^{\top}_j / \sqrt{\bm{d}_k}\right )}{\sum^T_{j=1}\textrm{exp}\left ( \bm{Q}_i\bm{K}^{\top}_j /\sqrt{\bm{d}_k}  \right )},
    \end{aligned}
\end{equation}
where $\bm{d}_k$ is the dimension of the keys.  Finally, the output of the self-attention-based neural network, which takes into account the importance of each input value, is given by $\textit{\textrm{Attention}}\left ( \bm{Q}, \bm{K},\bm{V} \right )=\bm{\alpha}\bm{V}$.

To enable the model to capture features of various patterns, multi-head attention is also well-suited for AD. The calculation of multiple heads is expressed as 
\begin{equation}
    \begin{aligned}
\textit{\textrm{Multihead}}\left (\bm{Q}, \bm{K}, \bm{V} \right ) = \textit{\textrm{Concat}}\left ( \textrm{head}_1, \cdots, \textrm{head}_h\right )\bm{W}_O,
    \end{aligned}
\end{equation}
where each head is computed as $\textrm{head}_i = \textit{\textrm{Attention}} \left ( \bm{Q}\bm{W}_{Q_i}, \bm{K}\bm{W}_{K_i}, \bm{V}\bm{W}_{V_i} \right )$. Here, $\bm{W}_{Q_i}$, $\bm{W}_{K_i}$, and $\bm{W}_{V_i}$ are trainable parameters for different heads, and $\bm{W}_O$ is the linear transformation matrix for the output. \( \text{Concat}(\text{head}_1, \cdots, \text{head}_h) \) concatenates the outputs of all attention heads along the feature dimension. Attention-based methods can effectively capture long-term dependencies, improve computational efficiency, and enhance the interpretability of AD through visualized attention weight values. When applied to AD, differences in the distribution of attention weights between normal and anomalous time series can serve as the basis for AD. 

In the field of AD, particularly for time series data, there has been a growing number of studies proposing deep learning methods based on attention mechanisms. Autoencoders that combine convolution, LSTM, and self-attention mechanisms can better extract complex features from multivariate time series data and robustly detect anomalies in high noise conditions \cite{70}. The Transformer, as a well-known attention-based model, has demonstrated superior performance in unsupervised prediction-based time series AD compared to LSTM, as it can learn the dynamic patterns of sequential data through self-attention mechanisms \cite{60}. The Transformer-based AD utilizes attention-based sequence encoders for rapid inference, achieving an F1 score improvement of up to 17\% on public datasets and reducing training time by as much as 99\% compared to the baseline \cite{64}. Despite its outstanding capabilities, the Transformer still faces certain bottlenecks in AD. Attention-based methods are prone to overfitting when data is insufficient. The method in \cite{78} seamlessly integrates contrastive learning and GAN into the Transformer, utilizing data augmentation techniques and geometric distribution masking to expand the training data, thereby enhancing data diversity and improving accuracy by 9.28\%.

Attention mechanisms are also frequently applied in graph neural networks to jointly detect anomalies in time series data. Reference \cite{65} proposes a novel efficient Transformer model based on graph learning methods, employing two-stage adversarial training to train the AD model and utilizing prototypical networks to apply the model to anomaly classification. A contrastive time-frequency reconstruction network for unsupervised AD is used for AD and localization \cite{68}, where attention mechanisms and graph convolutional networks update the feature information of each time point, combining points with similar feature relationships to dilute the influence of anomalous points on normal points. Reference \cite{72} models the correlations between temporal variables using graph convolutional networks, while also using an attention-based reconstruction model to output the importance of time series data within each time window, achieving an average AD F1 score exceeding 0.96. For multimodal data, a multimodal graph attention network (M-GAT) and temporal convolutional networks are used to capture spatial-temporal correlations in multimodal time series and correlations between modalities \cite{73}, ultimately outputting anomaly scores through reconstruction or prediction. More details about the application of GNNs in AD will be elaborated in the next subsection.

In addition to GNNs, CNNs can also incorporate attention mechanisms to enhance various metrics of AD. Reference \cite{80} effectively captures the local features of subsequences by leveraging the locality of CNNs and combining it with positional embeddings. At the same time, 
Zhu \textit{et al.}  \cite{80} employ attention mechanisms to extract global features from the entire time series, thereby enhancing the effectiveness and potential of detection. Many works have also introduced LSTM to extract temporal correlations in time series data based on CNN models with attention mechanisms. For example, Sun \textit{et al.} \cite{108} employ a sequential approach where 1D convolution is first used to extract abstract features of the signal values at each time step, which are then input into a bidirectional long short-term memory network (Bi-LSTM), ultimately combining with attention mechanisms to make the model focus on locally important time steps. Meanwhile, Le \textit{et al.} \cite{71} integrate convolutional layers, LSTM layers, and self-attention layers into an autoencoder architecture to better extract complex features from multivariate time series. 
Similarly, Pei \textit{et al.} \cite{9061946} employ additional SVM to classify the attention weights based on a CNN-LSTM model with attention mechanisms to determine whether cyber-attacks have occurred in energy systems. The input data are the multimodal measurements from the deployed sensors.
\subsubsection{GNN-based Anomaly Detection}
Graph Neural Networks (GNNs) have gained increasing attention in AD tasks, as many types of data can be naturally represented as graph structures \cite{jin2024survey}. Wu \textit{et al.} \cite{9471816} have demonstrated the effectiveness of GNNs in identifying anomalies within complex graph-structured data environments. As neural network models specifically designed to handle graph-structured data, GNNs define nodes, edges, and graphs, where nodes represent individual elements in the dataset, such as data points in a sequence, sensor readings in multivariate data, or entities in relational datasets—denoted as the set $\bm{V}$. Edges capture the relationships or dependencies between these elements, denoted as the set $\bm{E}$, and can represent temporal correlations, spatial dependencies, or more abstract relational connections depending on the context. The graph, represented as $\bm{G} = \left (  \bm{V}, \bm{E}\right )$, captures the overall structure formed by nodes and edges. The primary operations in GNN training are message passing and aggregation, which are used to update and learn node features. Specifically, during message passing, each node receives information from its neighboring nodes and updates its own state. For a node $v$,  the message passing formula is given as
\begin{equation}
    \begin{aligned}
\bm{m}^{(k)}_v = \sum_{u\in \mathcal{N}(v)} \textit{\textrm{MSG}}\left (\bm{h}^{(k-1)}_u, \bm{h}^{(k-1)}_v, \bm{e}_{uv} \right ),
    \end{aligned}
\end{equation}
where $\mathcal{N}(v)$ denotes the set of neighboring nodes of $v$, $\bm{h}_u$ and $\bm{h}_v$ are the features of nodes $u$ and $v$ at layer $k$, and $\bm{e}_{uv}$ represents the edge features. Subsequently, the received messages are aggregated with the current node state, and the node features are updated as
\begin{equation}
    \begin{aligned}
\bm{h}_v^{(k)} = \textit{\textrm{UPDATE}} \left (  \bm{h}_v^{(k-1)}, \bm{m}^{(k)}_v \right ),
    \end{aligned}
\end{equation}
where $\textit{\textrm{UPDATE}}(\cdot, \cdot)$ is the update function. 

\begin{figure}[!h]
	\vspace{-0.2cm}
	\centerline{\includegraphics[width=0.49\textwidth]{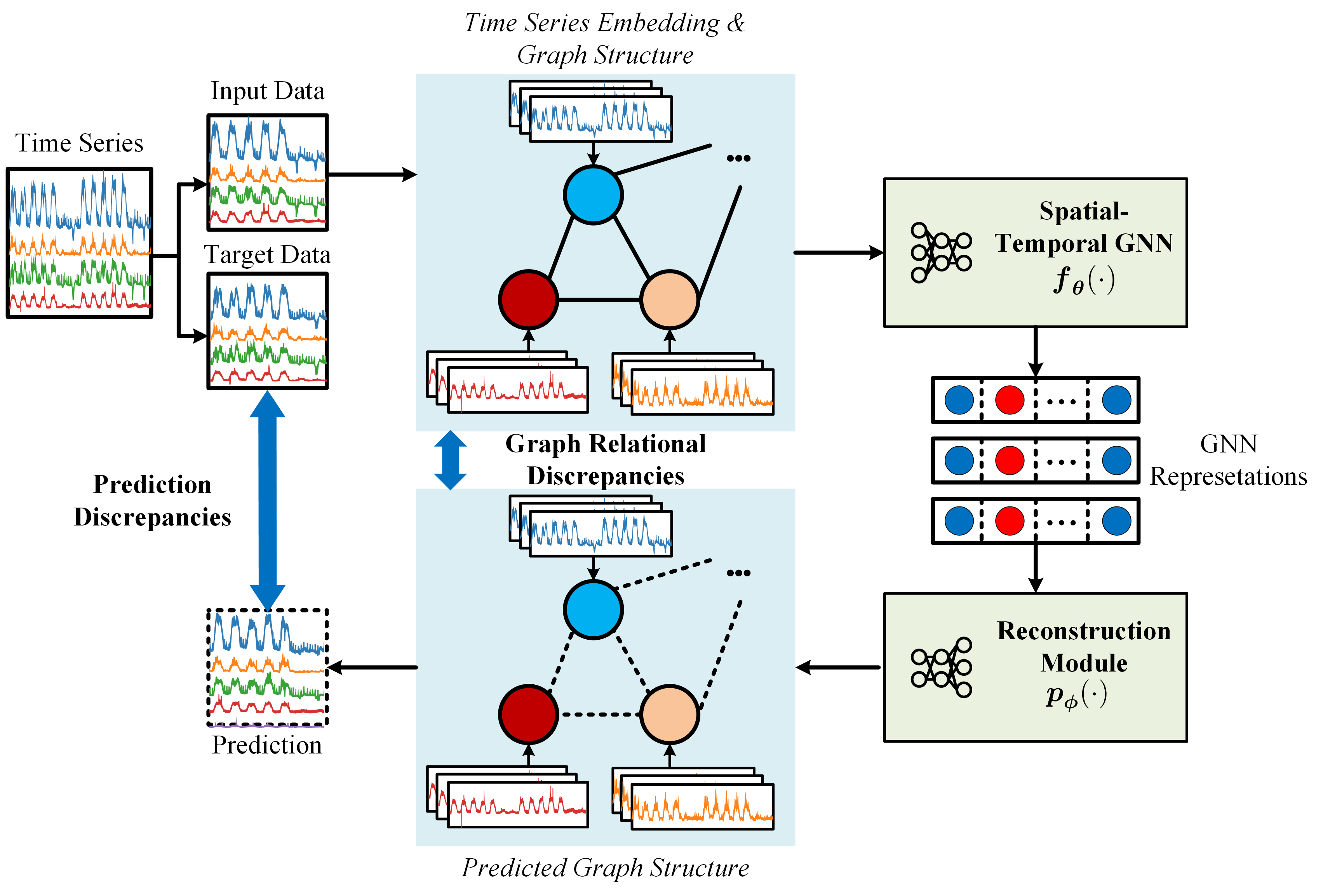}}
	\caption{GNN-based method for anomaly detection with time series data. Time series data is embedded into a graph structure, where a spatial-temporal GNN extracts dependencies. The reconstruction module then estimates the original data. Anomalies are detected based on graph relational discrepancies (differences in predicted graph structure) and prediction discrepancies (differences between reconstructed and actual time series).}
	\label{GNN_for_TSAD}
	\vspace{-0.2cm}
\end{figure}

As illustrated in Fig. \ref{GNN_for_TSAD}, which uses time series data as an example, GNNs treat each variable in the multivariate time series as a node to capture complex relationships between different dimensions. While the primary focus here is on the predictive capabilities of GNNs, it is worth noting that they are also effective in reconstruction-based AD. The final decision on whether the input sequence is anomalous is primarily based on prediction errors or graph structure differences, with reconstruction errors serving as a supplementary indicator. GNN-based AD methods excel at modeling complex dependencies between time steps or sensors, offering flexibility to handle both static and dynamic relationships across diverse time series structures. However, they still face challenges such as high computational complexity on large-scale graphs and difficulties in constructing optimal edge and graph configurations \cite{9906987}.

In prediction-based GNN for AD, GDN \cite{deng2021graph} is a representative work that combines a structure learning approach with GNN, additionally using attention weights to predict time series values and detect anomalies based on the predictions. Similar methods include GTA \cite{chen2021learning} and CST-GL \cite{zheng2023correlation}. Furthermore, Liu \textit{et al.} \cite{9395172} propose a GNN-based contrastive learning model that generates prediction scores from high-dimensional attributes and local structures to detect anomalies, outperforming state-of-the-art methods on seven benchmark datasets. Beyond prediction-based methods, there are also reconstruction-based GNN approaches. For example, MTAD-GAT \cite{zhao2020multivariate} employs a graph attention network as a spatiotemporal encoder to learn dependencies across variables and time, reconstructing the time series with a backbone reconstructor and identifying anomalies based on reconstruction errors. Similar techniques include VGCRN \cite{chen2022deep} and FuSAGNet \cite{han2022learning}.

\subsection{Deep learning methods for Anomaly Detection based on Hybrid Method}
In AD, reconstruction-based and prediction-based methods offer distinct but complementary approaches to identifying anomalies. Both methods rely on the discrepancy between the model's output and the actual input data as an indicator of abnormality. However, they diverge in how they handle data and their areas of application. Reconstruction-based methods focus on learning the underlying distribution of normal data. Once trained, the model attempts to recreate the input data. The reconstruction error, measured as the difference between the original data and its reconstruction, serves as a key indicator of anomalies. A high reconstruction error suggests that the data deviates from the normal patterns learned by the model. This approach is particularly effective in cases where understanding the full structure or distribution of the data is crucial, such as in image-based AD or other high-dimensional datasets. In contrast, prediction-based methods focus on forecasting specific attributes or missing values from the data, rather than reconstructing the entire input. These methods typically predict future values or infer missing data points by leveraging known features. If the predicted values significantly deviate from the actual values, this signals a potential anomaly. Prediction-based methods are often more suited to feature-rich datasets, where predicting specific variables can help identify irregular patterns. For instance, in applications like fraud detection, predicting expected behaviors or transactions can reveal anomalies when the predicted outcomes differ from the observed ones. While both methods differ in their data processing approaches, they can be highly complementary. In many cases, combining reconstruction-based and prediction-based techniques within a hybrid framework allows for more robust AD. Reconstruction models capture the overall structure and patterns in the data, while prediction models focus on detecting deviations in specific variables or features. This combination can provide a more comprehensive solution for identifying anomalies in complex datasets across various domains. Tang \textit{et al.} \cite{53} utilize a U-Net module as the prediction module to perform future frame prediction, amplifying reconstruction errors for abnormal events, while another U-Net module is used as the reconstruction module to enhance predicted frames for normal events, thus improving the effectiveness of AD. Lv \textit{et al.} \cite{75} adopt a dilated convolution-based autoencoder to integrate prediction errors and reconstruction errors into the output anomaly scores, effectively improving the generalization capability of the detection model. Liu \textit{et al.} \cite{54} leverage a reconstruction model and a prediction model within an end-to-end semi-supervised AD framework to effectively capture inter-variable correlations and temporal dependencies in multivariate time series data from wind turbines. Additionally, by incorporating an auxiliary discriminator with adversarial training, the model can progressively improve performance using limited labeled data, enhancing the transition from unsupervised to supervised AD. Wei \textit{et al.} \cite{116} propose a hybrid deep-learning model combining LSTM and autoencoder for AD in indoor air quality data, where the LSTM captures long-term dependencies in time-series data and the autoencoder uses reconstruction loss to detect anomalies, effectively addressing both temporal correlations and reconstruction errors for improved detection accuracy.

\subsection{Summary and Insights}
This section introduces three types of deep learning-based AD methods: reconstruction-based, prediction-based, and hybrid approaches. Reconstruction-based methods are particularly effective in handling high-dimensional and unsupervised data by learning intrinsic patterns and identifying deviations through reconstruction errors. Prediction-based methods excel at modeling temporal dependencies in time-series data, enabling the detection of unexpected patterns in dynamic environments. Hybrid approaches combine these strengths to address complex scenarios where multiple anomaly types coexist. Notably, these methods demonstrate the power of deep learning in capturing intricate patterns and dependencies that traditional methods often miss, making them indispensable for tackling diverse and challenging AD tasks.

\section{Integrate Traditional Method and Deep Learning Method} \label{sec:integrate}
In the field of AD, traditional methods and deep learning approaches each offer unique advantages. Traditional methods, such as clustering \cite{22} and Support Vector Data Description \cite{127}, are often simpler, more interpretable, and computationally efficient. These methods excel in providing transparent decision-making processes, making them suitable for applications where model interpretability is crucial. On the other hand, deep learning methods, with their ability to model complex, high-dimensional data distributions, offer enhanced detection accuracy and adaptability, especially for large datasets and unstructured data like images and sequences.

The integration of traditional and deep learning methods aims to leverage the interpretability and simplicity of traditional methods with the robustness and flexibility of deep learning techniques. By combining these approaches, researchers seek to create hybrid models that maintain accuracy while offering insights into the underlying decision-making process, improving both detection power and model transparency.
\subsection{Clustering method}
Clustering models play a crucial role in unsupervised AD, particularly for textual data. These models group similar data points based on their proximity in feature space and identify anomalies as points that deviate from established clusters \cite{26}. Common clustering techniques, such as k-means \cite{7}, Density-Based Spatial Clustering of Applications with Noise (DBSCAN) \cite{191}, and hierarchical clustering \cite{192}, work effectively for simpler datasets and offer the advantage of interpretability. By integrating clustering methods with deep learning, such as applying clustering post feature extraction by a neural network, it is possible to improve detection accuracy while maintaining an interpretable clustering structure. This hybrid approach is particularly useful in cases where data distribution varies, and flexible, context-aware AD is required. For instance, Li \textit{et al.} \cite{21} propose a method that extends fuzzy clustering with a reconstruction criterion and Particle Swarm Optimization (PSO) to detect anomalies in both amplitude and shape. This highlights how traditional clustering methods can benefit from optimization techniques to handle diverse anomaly types. Similarly, Markovitz \textit{et al.} \cite{123} introduce an innovative approach for AD in human actions by working directly on human pose graphs extracted from video sequences. By mapping these graphs to a latent space, clustering them, and applying a Dirichlet process-based mixture model, the method effectively leverages probabilistic modeling to enhance the robustness and flexibility of clustering for action recognition. In video AD, Qiu \textit{et al.} \cite{119} propose a convolution-enhanced self-attentive video auto-encoder integrated with a dual-scale clustering module based on the K-means algorithm. This approach effectively distinguishes normal and abnormal video data by enhancing feature representations and addressing the fuzzy boundaries between them. Additionally, Peng \textit{et al.} \cite{76} introduce a multivariate ELM-MI framework combined with a dynamic kernel selection method. By employing hierarchical clustering on unlabeled data to determine kernels, this method enables unsupervised online detection of various anomaly types, including point and group anomalies, while reducing computational costs and improving robustness. These studies collectively highlight the potential of hybrid approaches that integrate clustering with advanced techniques like deep learning, probabilistic modeling, or optimization frameworks. Such methods leverage the interpretability and simplicity of traditional clustering while addressing its limitations in handling complex data, offering a promising pathway for accurate and flexible AD.
\subsection{Normalizing Flows}
Normalizing Flows (NF) \cite{193} offer a probabilistic framework for AD by estimating the probability distribution of data. Using a sequence of invertible transformations, NFs can model complex distributions, making them particularly effective for identifying anomalies as low-probability events. When integrated with deep learning models, such as CNNs or RNNs, NFs act as precise probabilistic estimators, complementing the feature extraction capabilities of deep networks. This hybrid framework enhances AD, particularly in high-dimensional or unstructured datasets.

For instance, Yu \textit{et al.} \cite{29} propose FastFlow, a 2D normalizing flow module integrated with deep feature extractors like ResNet and Vision Transformers. By effectively modeling feature distributions and capturing both local and global relationships, FastFlow achieves state-of-the-art performance, with a 99.4\% AUC on the MVTec AD dataset, while maintaining high inference efficiency. Similarly, Cho \textit{et al.} \cite{30} introduce Implicit Two-path Autoencoder (ITAE), which reconstructs normal video patterns by implicitly modeling appearance and motion features through two encoders and a shared decoder. NF enhances ITAE by estimating the density of normal embeddings, enabling robust detection of out-of-distribution anomalies, with strong results across six surveillance benchmarks. For multivariate time series data, Zhou \textit{et al.} \cite{63} combine a graph structure learning model with entity-aware normalizing flows to capture interdependencies and evolving relations among entities. By estimating entity-specific densities and employing a clustering strategy for similar entities, the extended MTGFlow\_cluster improves density estimation accuracy, demonstrating superior performance on six benchmark datasets. Further expanding on the use of graphs, Dai \textit{et al.} \cite{155} propose Graph-Augmented Normalizing Flow (GANF), which incorporates a Bayesian network to model causal relationships among time series. This approach factorizes joint probabilities into conditional probabilities, improving density estimation and enabling effective detection of anomalies in low-density regions, as well as identifying distribution drifts.

These studies collectively highlight the strengths of integrating Normalizing Flows with traditional and deep learning-based methods. By combining the interpretability and precision of probabilistic models with the expressive power of deep networks or graph structures, these hybrid approaches address the challenges of complex data distributions, offering scalable and robust solutions for diverse AD tasks. This synergy underscores the potential of such methods to push the boundaries of accuracy and adaptability in real-world applications.
\subsection{Support Vector Data Description}
Support Vector Data Description (SVDD) \cite{127} is a traditional machine learning method used to define a boundary around normal data points, effectively distinguishing them from anomalies. Unlike binary classification, SVDD is particularly effective for one-class classification tasks, where only normal data is available. This approach is computationally efficient and interpretable, as it provides a clear boundary between normal and abnormal points. By integrating SVDD with deep learning, researchers can enhance the boundary definition based on high-dimensional features extracted by a neural network, resulting in a model that combines the boundary precision of SVDD with the feature richness of deep learning. This hybrid model is highly effective in scenarios where boundary clarity and interpretability are paramount, such as in industrial monitoring or fraud detection.

To improve latent representations, Zhou \textit{et al.} \cite{31} propose Deep SVDD-VAE, which jointly optimizes VAE and SVDD. The VAE reconstructs input data, and SVDD simultaneously defines a spherical boundary in the latent space, ensuring separability of normal and anomalous instances. This joint optimization significantly outperforms traditional AE-based methods, as shown on MNIST, CIFAR-10, and GTSRB datasets. For variable-length time series data, Ergen \textit{et al.} \cite{81} introduce an LSTM-based AD framework, where LSTM and SVDD are jointly optimized using modified objectives. This method extends seamlessly to GRU architectures, demonstrating strong performance across unsupervised, semisupervised, and supervised settings. Besides, Zhang \textit{et al.} \cite{126} propose Deep Structure Preservation SVDD (DSPSVDD), which simultaneously minimizes hypersphere volume and network reconstruction error. This dual objective ensures deep feature preservation and enhances AD performance, outperforming traditional SVDD models on datasets like MNIST and MVTec AD.

These studies highlight the strengths of combining SVDD with deep learning, where deep models enhance feature representation while SVDD ensures boundary precision. This hybrid framework effectively addresses limitations in both methods, offering a scalable and interpretable solution for complex AD tasks across diverse domains.

\subsection{Summary and Insights}
This section explores the integration of traditional and deep learning methods for AD, highlighting how their complementary strengths can be combined. Traditional methods, known for their simplicity, interpretability, and computational efficiency, excel in scenarios where transparency is critical. In contrast, deep learning methods offer superior adaptability and accuracy, particularly for high-dimensional and unstructured data. By integrating these approaches, hybrid models can leverage the interpretability of traditional methods while retaining the robustness and flexibility of deep learning. This fusion not only enhances AD performance but also bridges the gap between accuracy and model transparency, making it a promising direction for future research.

\section{Open issues and future works} \label{sec:open}
\subsection{Data Collection}
Data scarcity and class imbalance remain major challenges in AD. Since anomalies are rare, obtaining large labeled datasets is costly and time-consuming, especially when expert annotation is required. Supervised learning struggles due to the lack of abnormal samples, while the overwhelming presence of normal data biases models toward common patterns. This problem is particularly critical in cybersecurity, healthcare, and industrial monitoring, where undetected anomalies can have serious consequences.

Several approaches mitigate these issues. Semi-supervised and unsupervised learning exploit normal data distributions to detect deviations without requiring labeled anomalies \cite{luo2022smd} \cite{li2021cutpaste}. Data augmentation, synthetic data generation, and oversampling improve data balance by increasing the number of anomalous examples, helping models generalize better \cite{liu2023anomaly} \cite{wen2020time}. Despite these advancements, challenges remain. Semi-supervised methods struggle with subtle anomalies that closely resemble normal data. Augmentation techniques, often based on simple transformations, may fail to capture complex domain-specific variations. Similarly, synthetic data generation may not fully reflect real-world anomaly diversity, leading to models biased toward normal samples. Moreover, even with augmentation, models risk overfitting to the majority class, compromising anomaly detection performance. Ensuring that models remain sensitive to rare anomalies while maintaining accuracy on normal data remains an ongoing challenge. Future research may focus on refining self-supervised learning \cite{hojjati2024self}, improving the diversity of synthetic samples \cite{zhang2024realnet}, and developing more adaptive anomaly detection frameworks to enhance robustness in real-world applications.

\subsection{Computational Complexity}
In AD, computational complexity is a crucial factor, especially for systems operating in real-time environments or handling large-scale datasets. The efficiency of an algorithm directly impacts its feasibility in fields like industrial monitoring, cybersecurity, and autonomous systems, where swift detection is essential. Many advanced models, particularly deep learning approaches like autoencoders, GANs, and LSTMs, are computationally intensive due to their complex architectures and iterative learning processes. This often leads to trade-offs between detection accuracy and computational efficiency, with continuous efforts aimed at optimizing models to reduce computational demands without sacrificing performance.

Moreover, AD models frequently require substantial memory resources, especially when dealing with high-dimensional or streaming data, making memory usage a crucial consideration. Techniques like memory-efficient architectures, data compression, and sparse modeling are commonly used to address this issue. Real-time AD adds further complexity, as algorithms must process incoming data and make rapid decisions in applications like autonomous driving and fraud detection \cite{van2019real}, where even minimal delays can have severe consequences. Achieving real-time performance typically involves optimizing data processing speeds and decision-making through lightweight models \cite{abououf2022self} \cite{zhou2024reconstructed} and parallel processing techniques, such as GPU acceleration \cite{zhao2021tod}. However, balancing real-time detection capabilities with high accuracy remains challenging.

The tension between computational complexity and detection accuracy persists, as complex models often excel in detection but lack practical applicability for real-time or large-scale scenarios. Simpler models, though computationally efficient, may fail to detect nuanced anomalies. Hybrid models or multi-stage frameworks that deploy complex methods only as needed provide a potential solution. Additionally, future research may benefit from exploring distributed computing solutions, like cloud \cite{al2024anomaly} or edge computing, to enhance real-time AD performance in resource-limited environments.

\subsection{Explainability and Interpretability}
Deep learning methods have greatly advanced AD by capturing complex patterns in high-dimensional data. However, they are often criticized as ``black-box" models due to their lack of transparency, making it challenging to understand why certain data points are flagged as anomalies. For fields like healthcare, finance, or industrial monitoring, accurate detection alone is insufficient; stakeholders also need clear explanations to understand why a particular anomaly was detected. This lack of interpretability limits the practical deployment of deep learning models, as the inability to justify decisions reduces trust and hinders adoption in critical applications.

In fields like healthcare, where anomalies may be linked to medical diagnoses, or in finance, where fraud detection can carry legal implications, interpretability is essential. Transparent model decisions enable experts to validate results and make informed decisions. In safety-critical applications, such as autonomous driving or industrial equipment monitoring, understanding the rationale behind AD is vital for ensuring safety. One major challenge is balancing the trade-off between model interpretability and performance. Simpler models, like decision trees or linear regression, offer greater transparency but often lack the complexity needed to detect subtle anomalies in high-dimensional data. In contrast, deep learning models provide high accuracy but are harder to interpret.

Ongoing research is exploring hybrid approaches, where interpretable models are combined with more complex ones, allowing for accurate AD with the added benefit of interpretability. For example, attention mechanisms \cite{151} in neural networks can help highlight specific data regions influencing decisions, providing insights into the model’s internal workings. Alternatively, tools like Local Interpretable Model-agnostic Explanations (LIME) and SHapley Additive exPlanations (SHAP) \cite{194} can offer post-hoc explanations, improving transparency without altering model structure. Future research could also focus on real-time explainability in time-sensitive applications, and incorporating domain knowledge or user feedback to enhance model interpretability.

\subsection{Handling Diverse Types of Anomalies}
In real-world AD, multiple types of anomalies often coexist, adding complexity to the detection process. Beyond point anomalies, which are the simplest, other types like contextual and collective anomalies are common, especially in dynamic environments. For instance, in intelligent transportation systems, anomalies may include both isolated incidents (e.g., a single vehicle’s sudden deceleration) and collective patterns (e.g., multiple vehicles simultaneously slowing down), each requiring different detection methods. Effectively capturing these varied anomaly types requires flexible models capable of adapting to different anomaly patterns without focusing on only one type.

Continuous research is needed to develop models that can generalize across anomaly types, enhancing adaptability and balancing detection accuracy with model flexibility. Hybrid approaches, for instance, can integrate different methods to capture diverse anomalies more effectively. The challenge remains in achieving this versatility without sacrificing accuracy, as models must maintain strong performance across different contexts. Future work may also explore multi-modal models \cite{liu2024uac} that combine different types of data, further improving detection capabilities by drawing from diverse data sources. These directions aim to create AD systems that are both robust and adaptable, capable of handling the complex and mixed nature of real-world anomaly scenarios.

\section{Conclusion} \label{sec:con}
In this survey, we have provided a comprehensive overview of the recent advancements in AD with a primary focus on deep learning techniques from 2019 to 2024. By analyzing over 180 research papers from leading journals and conferences, we have explored how AD methods have evolved to address diverse challenges across various types of data. This survey categorizes and examines deep learning methods into reconstruction-based, prediction-based, and hybrid approaches, highlighting their strengths, limitations, and applications. Recognizing the simplicity, interpretability, and computational efficiency of traditional AD methods, we reviewed their integration with deep learning techniques. These hybrid approaches aim to leverage the strengths of both paradigms, enhancing robustness and efficiency in AD systems. This survey not only sheds light on the state-of-the-art techniques but also identifies gaps and opportunities for future research. By focusing on the latest trends and innovations, this work aims to inspire further exploration and advancements in the rapidly evolving field of AD.


%


\small
\bibliographystyle{IEEEtran}
\bibliography{reference}{}

\vfill

\end{document}